\documentclass{article}

\usepackage{PRIMEarxiv}

\usepackage{enumitem}
\usepackage{amsmath}
\usepackage{mathtools}
\usepackage{amsthm}
\usepackage{longtable}
\usepackage{comment}
\usepackage{indentfirst}

\usepackage{natbib}

\usepackage[utf8]{inputenc}
\usepackage[T1]{fontenc}
\usepackage{hyperref}
\usepackage{url}
\usepackage{booktabs}
\usepackage{amsfonts}
\usepackage{amssymb}
\usepackage{nicefrac}
\usepackage{microtype}
\usepackage{graphicx}
\usepackage{multirow}
\usepackage{array}
\usepackage{siunitx}
\usepackage{xcolor}
\usepackage{listings}
\usepackage{graphics}
\usepackage{subcaption}
\usepackage{caption}
\usepackage{float}
\usepackage{pdflscape,adjustbox}
\usepackage{makecell}
\usepackage[table]{xcolor}

\definecolor{codegreen}{rgb}{0,0.6,0}
\definecolor{codegray}{rgb}{0.5,0.5,0.5}
\definecolor{codepurple}{rgb}{0.58,0,0.82}
\definecolor{backcolour}{rgb}{0.97,0.97,0.97}

\title{\includegraphics[width=\textwidth]{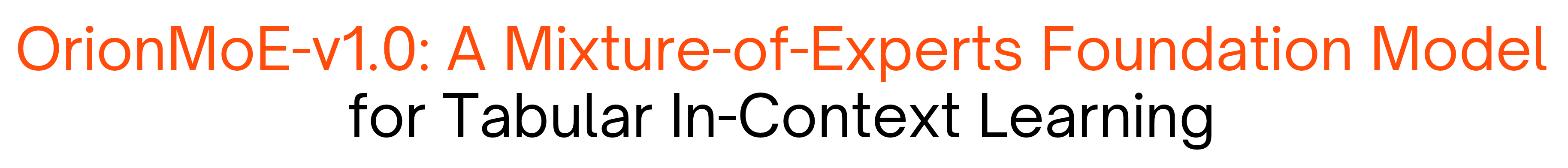}}

\author{
  Mohamed Bouadi$^*$, Nassim Bouarour$^*$,\\
  Aditya Tanna, Vinay Kumar Sankarapu \\
  \affiliation{Lexsi Labs}\\
   \small{$^*$Equal contribution}
}

\setabstract{%
Tabular foundation models enable in-context learning by predicting unseen observations from a labeled support table without task-specific parameter updates. Despite recent advances, most existing approaches rely on dense architectures that apply the same transformations across features and datasets. We introduce \textsc{OrionMoE-Small-v1.0}, a compact tabular foundation model that combines distribution-aware representations with dataset-conditioned sparse mixture-of-experts computation. Built on a dual-axis transformer architecture, \textsc{OrionMoE-Small} selectively activates experts with routing decisions informed by both individual cell representations and the broader dataset context. This design enables adaptive allocation of computation across heterogeneous tabular problems while retaining a shared computational pathway across tasks. The model is pretrained entirely on synthetic classification tasks generated from a structured causal prior and performs downstream prediction without gradient-based adaptation. Experiments across multiple tabular benchmarks demonstrate competitive performance against substantially larger foundation models, with particularly strong results on small-context and high-dimensional datasets. Further analysis reveals favorable relative performance under class imbalance and highlights the importance of feature composition.}

\setkeywords{Tabular Foundation Model, Mixture-of-Experts, In-Context Learning, Tabular Classification}

\setlicense{Lexsi Labs Source Available License (LSAL) v1.2, OrionMoE Edition (see Section \ref{sec:license})}

\runningtitle{OrionMoE-Small-v1.0: A Mixture-of-Experts Foundation Model for Tabular In-Context Learning}

\begin{document}
\maketitle

\section{Introduction}
\label{sec:introduction}

\begin{table*}[t]
\centering
\small
\setlength{\tabcolsep}{8pt}
\caption{
Model capacity and predictive performance of recent tabular foundation models. Models are grouped into three capacity regimes: small ($<50$M parameters), medium ($50$--$150$M), and large ($>150$M). We report Elo ratings on TabArena ($N=38$), BeyondArena ($N=98$), OpenML-CC18 ($N=48$), and TALENT ($N=148$). Higher Elo indicates better predictive performance. Ratings are computed independently within each benchmark and are not directly comparable across benchmarks. A dash indicates unavailable results.
}
\label{tab:teaser}
\begin{tabular}{lccccc}
\toprule
\textbf{Model}
& \textbf{Params (M)}
& \textbf{TabArena}
& \textbf{BeyondArena}
& \textbf{OpenML-CC18}
& \textbf{TALENT} \\
\midrule

\multicolumn{6}{l}{\textit{Small models ($<50$M)}} \\
\rowcolor{gray!12}
\textbf{OrionMoE-Small-v1.0} & 37.7 & 1320 & 1149 & 1118 & 1024 \\

Causilo & 36.1 & 1530 & 1268 & 1170 & 1125 \\

TabICLv2 & 27.6 & 1277 & 1089 & 1121 & 1054 \\

Kumo-Tabular-Small & 28 & 1558 & -- & 1241 & 1150 \\

\midrule
\multicolumn{6}{l}{\textit{Medium models ($50$--$150$M)}} \\
TabPFN-3 & 53 & 1351 & 1166 & 1015 & 1021 \\

TabDPT-1.3 & 63 & 1215 & 1041 & 1061 & 974 \\

Xiaomi-TabLDM & 70 & 1287 & -- & 1083 & 1002 \\

Kumo-Tabular-Medium & 62 & 1726 & -- & 1334 & 1265 \\

\midrule
\multicolumn{6}{l}{\textit{Large models ($>150$M)}} \\
Kumo-Tabular-Large & 215 & \textbf{1795} & -- & \textbf{1447} & \textbf{1338} \\

TabPFN-3.5 & 220 & 1677 & \textbf{1348} & 1231 & 1231 \\

LimiX-2 & 400 & 1769 & 1188 & 1387 & -- \\

TabFM & 408.7 & 1583 & 1222 & 1372 & -- \\

\bottomrule
\end{tabular}%
\end{table*}

Tabular data remains one of the most prevalent forms of structured information in machine learning, underpinning applications in healthcare, finance, scientific discovery, and industrial systems \cite{henry2015targeted,khandani2010consumer,carvalho2019systematic,baldi2014searching}. Unlike images or natural language, however, tabular datasets exhibit substantial heterogeneity in their statistical structure. Features may represent continuous measurements, categorical attributes, counts, or ordinal quantities, with widely varying distributions, scales, and predictive relationships. Dataset sizes, dimensionality, missingness, and label distributions also differ considerably across applications. This diversity has historically favored specialized learning algorithms, particularly gradient-boosted decision trees, over neural architectures that must be trained separately for each task \citep{gorishniy2021revisiting,grinsztajn2022tree,reichstein2019deep,wang2023scientific,shwartz2022tabular}.

Recent advances in tabular foundation models (TFMs) have begun to reshape this landscape. Models such as TabPFN \citep{tabpfn,tabpfnv2,tabpfnv3,tabpfn-3.5}, TabICL \citep{tabicl,tabiclv2}, OrionMSP \citep{orionmsp}, OrionBix \cite{orionbix}, LimiX \cite{limix,limix-2}, TabDPT \cite{tabdpt}, EXAONE Tabular \cite{exaone} and TabFM \citep{tabfm} demonstrate that a single pretrained network can transfer predictive capabilities across diverse tabular datasets. A central paradigm underlying these models is \emph{in-context learning}: given a labeled support table, the model predicts unseen observations through a forward pass, without task-specific parameter updates. By pretraining on large collections of synthetic prediction problems, TFMs amortize the cost of learning across tasks, replacing repeated optimization with contextual adaptation at inference time. Recent evaluations, including TabTune \citep{tanna2025tabtune} and TabArena \citep{tabarena}, highlight the increasing competitiveness of this approach against conventional tabular learning methods.

Despite this progress, an important architectural question remains relatively unexplored: \emph{should every feature and dataset be processed by the same computational pathway?} Existing in-context TFMs predominantly rely on dense transformer-style architectures, where feed-forward transformations are shared across feature representations and datasets. Although attention mechanisms allow these models to adapt their representations to the observed context, the underlying feed-forward computation remains dense and uniformly applied. Yet the statistical mechanisms governing tabular features can differ substantially, even within a single dataset. A continuous measurement, a high-cardinality categorical attribute, and a discrete variable may require different transformations to extract predictive information. Moreover, the relevance of a particular transformation may depend on the broader dataset context rather than on the individual feature alone. This raises the possibility that adapting the \emph{allocation of computation}, in addition to the representations themselves, could provide a useful inductive bias for tabular foundation models.

Mixture-of-Experts (MoE) architectures offer a natural framework for conditional computation. By selecting a subset of specialized networks for each input, sparse MoE models increase the diversity of available transformations without requiring every expert to be evaluated at every step \citep{shazeer2017outrageously,fedus2022switch}. This principle has proved effective in scaling neural models, particularly in language modeling, where learned routing mechanisms allocate computation across different input representations. Its application to tabular in-context learning, however, raises a distinct challenge. Unlike language tokens, tabular feature representations have no universal semantics across datasets: the same numerical value may correspond to entirely different quantities, distributions, and predictive relationships. Consequently, expert selection should account not only for the local representation of a feature but also for the statistical context of the dataset in which it appears.

In this report, we introduce \textsc{OrionMoE-Small-v1.0}, a compact tabular foundation model that combines in-context learning with dataset-conditioned sparse computation. \textsc{OrionMoE-Small} retains the two-dimensional structure of tabular data through a dual-axis backbone that separates information exchange across observations from interactions among features and samples. Its central innovation is to replace dense feed-forward computation with a sparse mixture of experts. Each token is processed by a small subset of routed experts, allowing the model to select among multiple learned cell transformations while preserving a common computational pathway.


\textsc{OrionMoE-Small-v1.0} is pretrained entirely on synthetic prediction problems generated using an extended version of \textsc{O'PRIOR} \cite{oprior}, our synthetic task generator. The prior combines diverse structural causal mechanisms with heterogeneous feature distributions and controlled variations in dataset characteristics. This provides a broad family of learning problems in which different features and datasets may exhibit distinct statistical relationships, offering a natural setting for learning conditional feature transformations and dataset-aware routing.

We evaluate \textsc{OrionMoE-Small-v1.0} against recent tabular foundation models across multiple benchmark suites, including TabArena \cite{tabarena}, OpenML-CC18 \cite{openmlcc18}, TALENT \cite{talent}, and BCCO \cite{limix}.  Table~\ref{tab:teaser} provides an overview of the relationship between model capacity and predictive performance across recent tabular foundation models. To contextualize \textsc{OrionMoE-Small-v1.0}, we distinguish three parameter count regimes: small ($<50$M), medium ($50$--$150$M), and large ($>150$M). With only $37.7$ million parameters, \textsc{OrionMoE-Small} belongs to the small-model category, alongside Causilo \cite{causilo} and TabICLv2 \cite{tabiclv2}. Despite its compact size, \textsc{OrionMoE-Small} achieves competitive performance across all four benchmarks, outperforming TabICLv2 on TabArena and BeyondArena and remaining close to the larger TabPFN-3 model. It also surpasses the medium-sized TabDPT-1.3 across all four benchmarks. Although the strongest large-scale models, particularly Kumo-Tabular, TabPFN-3.5 and LimiX-2, generally achieve higher predictive performance, their parameter budgets are substantially larger. At the same time, Causilo's strong performance within the small-model category indicates that model size alone does not explain predictive competitiveness. 




\section{Model Architecture and Inference}
\label{sec:model}

\textsc{OrionMoE-Small-v1.0} is a tabular foundation model that combines in-context learning with dataset-conditioned sparse computation. Given a labeled support set $\mathcal{S}=\{(\mathbf{x}_i,y_i)\}_{i=1}^{N}$ and a set of unlabeled query observations $\mathcal{Q}=\{\mathbf{x}_j^*\}_{j=1}^{M}$, the model predicts the query targets by conditioning on $\mathcal{S}$, without task-specific parameter updates. Adaptation to a new dataset therefore occurs entirely through the forward pass.

The architecture is motivated by two properties of tabular data. First, the meaning of an individual feature value depends on the empirical distribution and predictive relationships of the dataset in which it appears. Second, different features and datasets may require distinct computations to extract predictive information. \textsc{OrionMoE-Small-v1.0} addresses these properties through a distribution-aware column encoder introduced in TabICLv2 \cite{tabiclv2}, a dual-axis transformer backbone, and a sparse mixture-of-experts (MoE) layer whose routing decisions are conditioned on a learned dataset representation. An in-context prediction head subsequently aggregates the learned representations to produce task-conditioned predictions.

\subsection{Model Architecture}
\label{sec:model_architecture}

\paragraph{Distribution-aware column embedding.}
\textsc{OrionMoE-Small} represents a table as a grid of cell embeddings, retaining a separate representation for each observation--feature pair. Following TabICLv2 \cite{tabiclv2}, a shared column encoder contextualizes each value using the support observations in its column. This provides a data-dependent representation of each feature without relying on fixed column identities.

Specifically, scalar values are first projected into the model's latent space and processed by a sequence of induced self-attention blocks. In each block, learned inducing vectors aggregate information from the support observations, and both support and query cells attend to the resulting column summaries. A cell's representation therefore depends on its value and on the statistical context supplied by the support column: the same numerical value can receive different embeddings in different datasets.

The column encoder is also target-aware. Support-target embeddings are added to the corresponding support representations before induced attention, allowing the column summaries to capture information relevant to the prediction task. Query targets are never provided, and query observations do not contribute to the construction of the support summaries. We adapt this encoding scheme by processing columns individually, leaving cross-feature interactions to the subsequent dual-axis backbone. The contextual cell embeddings are combined with a learned missingness signal, preserving the distinction between absent entries and observed numerical values.

In addition to the original feature tokens, \textsc{OrionMoE-Small-v1.0} introduces learned feature registers. These registers participate in within-row computation and serve as dedicated representations for collecting information across the features of each observation. 

\paragraph{Dual-axis representation learning.}
A central design principle of \textsc{OrionMoE-Small-v1.0} is to preserve the two-dimensional structure of tabular data. Each backbone block combines computation along the sample and feature axes, allowing representations to capture complementary dependencies.

Along the row axis, attention operates across observations within each feature. This enables a cell representation to incorporate information from other observations belonging to the same dataset. Along the feature axis, attention operates across feature tokens within each observation, modeling dependencies among its attributes. This structure allows statistical information aggregated across observations and interactions among features to inform one another. Repeating these operations progressively integrates information across both axes while preserving individual feature representations.

To support large in-context datasets, row-axis attention uses an induced set attention mechanism. A fixed number of learned inducing vectors first aggregate information from the support observations, and individual observations then attend to the resulting latent summaries. With a fixed number of inducing vectors, the row-axis attention cost scales linearly with the number of observations, rather than quadratically as in conventional full self-attention.

The attention mechanism also enforces the in-context information boundary: support representations are computed exclusively from the labeled support set, while query representations attend only to support-derived information. Consequently, query observations cannot alter the support state used to predict other queries.

Feature-axis attention, by contrast, operates directly over the feature tokens and learned registers. Together, the two attention mechanisms separate the aggregation of evidence across observations from the modeling of interactions among features.

\paragraph{Sparse mixture-of-experts computation.}
The principal architectural contribution of \textsc{OrionMoE-Small-v1.0} is the introduction of sparse conditional computation. Conventional dense tabular transformers apply the same feed forward transformation to every feature token, irrespective of the statistical properties of the feature or the dataset. \textsc{OrionMoE-Small} instead equips each block with a collection of expert networks and dynamically selects which transformations to apply.

For a representation $\mathbf{h}$ and a dataset representation $\mathbf{r}$, the MoE transformation is defined as
\begin{equation}
\operatorname{MoE}(\mathbf{h},\mathbf{r}) = \sum_{e \in \mathcal{T}_k(\mathbf{h},\mathbf{r})} \alpha_e(\mathbf{h},\mathbf{r}) E_e(\mathbf{h}),
\label{eq:orion_moe}
\end{equation}
where $E_e$ represents routed expert $e$, and $\mathcal{T}_k$ contains the indices of the $k$ experts with the highest routing scores. The corresponding routing weights are normalized over the selected experts.

\paragraph{Dataset-conditioned expert routing.}
Beyond token-dependent expert selection, \textsc{OrionMoE-Small-v1.0} introduces dataset-conditioned routing. The underlying motivation is that the appropriate transformation of a token may depend not only on its local representation, but also on the broader statistical structure of the table.

To capture this information, the model maintains learned dataset registers $\mathbf{r}$, which are progressively updated through attention-based aggregation over support representations. A register summarizes information about the current dataset and influences expert selection throughout the backbone. 

This mechanism distinguishes \textsc{OrionMoE-Small-v1.0} from conventional token-only sparse routing. Two tokens with similar local representations may be assigned to different experts when they occur in different datasets. Conversely, the router may identify common computational requirements across different datasets and direct them to the same experts. Importantly, expert specialization is learned rather than prescribed: every routed expert remains eligible for every token, and routing is not constrained by feature type, dataset identity, or application domain. 

\paragraph{In-context prediction head.}
Following the dual-axis backbone, the feature registers are concatenated to obtain a fixed-width representation of each observation, which is combined with a projected representation of dataset registers.  A stack of in-context attention layers then processes the resulting support and query representations. Support targets are incorporated into the context, allowing query predictions to depend on the relationships inferred from the labeled observations.

For classification, a decoder maps the resulting query representations to class logits. Our model supports a native budget of ten classes and extends to larger label spaces through a hierarchical classification procedure. 
The resulting architecture combines distribution-aware cell representations, structured information exchange across the two axes of a table, and dataset-conditioned conditional computation within a single in-context prediction framework.

At inference time, \textsc{OrionMoE-Small-v1.0} adapts to a new dataset solely by conditioning on its labeled support table. The pretrained parameters remain fixed, and no task-specific optimization is required. The support/query separation also enables efficient reuse of support-derived representations across successive query batches. Since query observations cannot modify the support state, queries may be processed in chunks against a cached context without changing the resulting predictions, provided the support set and inference configuration are unchanged.

\subsection{Synthetic Prior Generation}
\label{sec:synthetic_prior}

\textsc{OrionMoE-Small-v1.0} is pretrained on synthetic in-context learning episodes generated using \textsc{O'PRIOR} \cite{bouadiyou}, our previously released synthetic task generator. \textsc{O'PRIOR} provides a diverse collection of structural causal model families, including multilayer perceptrons, tree-based mechanisms, convolutions, Gaussian processes, time-lagged models, linear functions, and hybrid compositions. Together, these mechanisms capture a broad range of statistical relationships, from smooth nonlinear dependencies to discontinuous decision boundaries and temporal interactions. By varying the underlying structures, functional mechanisms, and sampling parameters, the prior generates prediction tasks spanning different dimensionalities, sample sizes, and levels of predictive complexity.


Beyond the underlying causal mechanisms, \textsc{O'PRIOR} supports composable transformations that introduce heterogeneous feature distributions, redundant attributes, outliers, different target distributions, and structured missingness. The generator also provides optional mechanisms for latent confounding, spurious correlations, and support--query distribution shifts. Transformations requiring fitted statistics are estimated exclusively from support observations, preserving the information boundary between support and query sets. For \textsc{OrionMoE-Small-v1.0}, we additionally introduce controlled count-, ordinal-, and proportion-valued features, together with stage-dependent missingness patterns, to broaden the statistical conditions encountered during pretraining.

The resulting prior exposes \textsc{OrionMoE-Small-v1.0} to variation both within and across datasets. Heterogeneous generative mechanisms create opportunities for distinct feature-level transformations, while variations in dataset structure and statistical properties provide contextual information that can influence expert routing. This diversity is aligned with \textsc{OrionMoE-Small}'s dataset-conditioned mixture-of-experts architecture, which learns to allocate computation based on both individual feature representations and the broader dataset context.

\section{Experimental Results}

\subsection{TabArena}
\label{sec:tabarena}

TabArena~\citep{tabarena} provides a competitive evaluation of tabular learning methods across diverse classification tasks, including recent foundation models, conventional machine learning algorithms, and tuned or ensembled systems. We evaluate \textsc{OrionMoE-Small-v1.0} on 38 classification datasets, considering both overall predictive performance and its relationship with computational cost. We further examine performance across task types, dataset sizes, and target-class distributions to characterize the regimes in which \textsc{OrionMoE-Small-v1.0} is most competitive.

\begin{figure}[htbp]
    \centering
    \includegraphics[width=\linewidth]{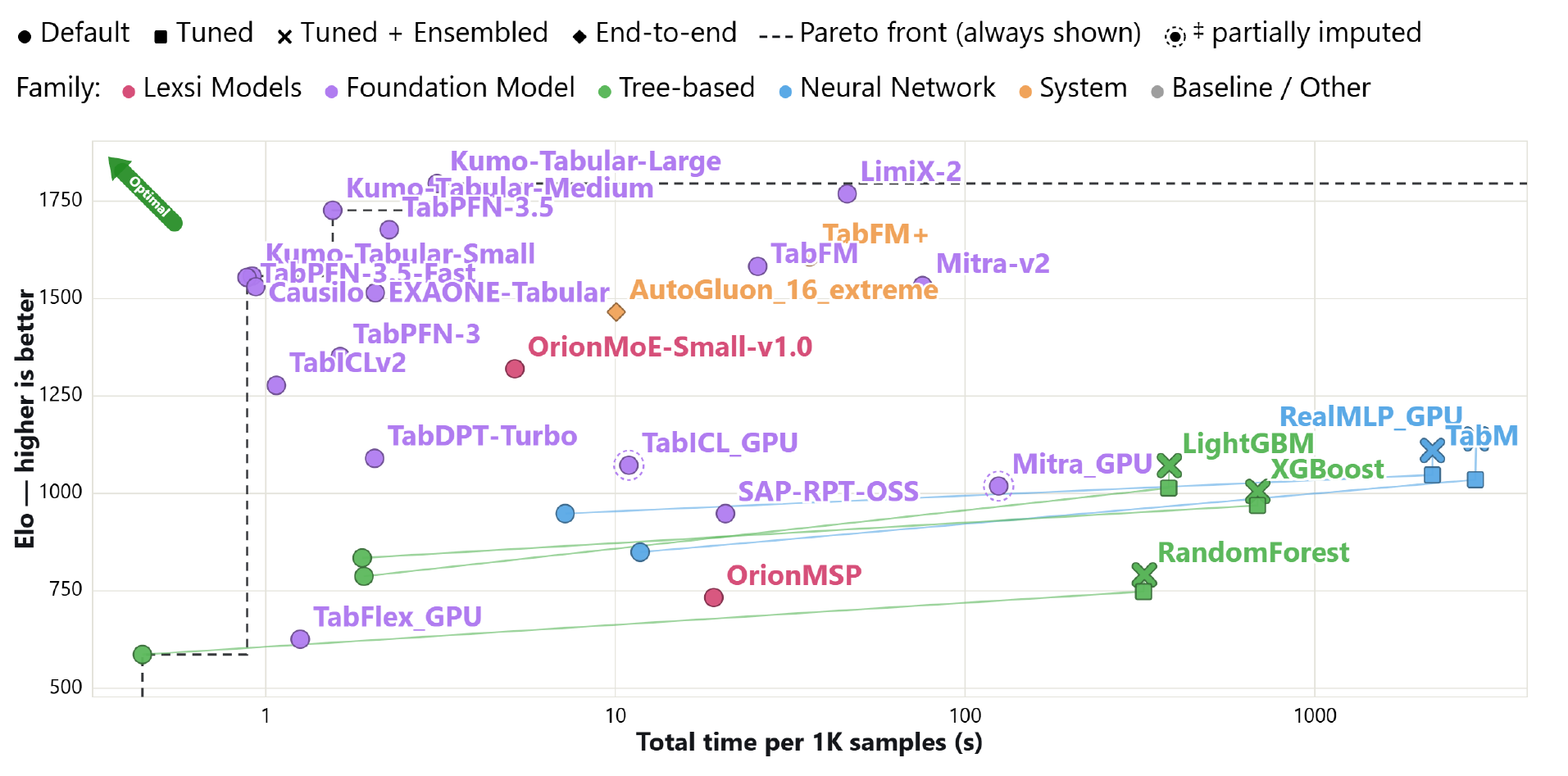}
    \caption{Elo against total runtime per 1K samples on 38 TabArena classification tasks. Total time is the fit time per 1,000 training rows plus the prediction time per 1,000 test rows, median over the 38 classification datasets.}
    \label{fig:elo-total-tabarena}
\end{figure}

Figure~\ref{fig:elo-total-tabarena} presents the trade-off between Elo rating and total runtime, including support-set processing and prediction time. \textsc{OrionMoE-Small-v1.0} obtains an Elo rating of $1320$, placing it ahead of several established tabular foundation models, including TabICLv2 ($1277$) \cite{tabiclv2}, Xiaomi-TabLDM ($1287$) \cite{tabldm}, and TabDPT-1.3 ($1215$) \cite{tabdpt}. \textsc{OrionMoE-Small-v1.0} remains close to TabPFN-3 ($1351$) \cite{tabpfnv3}, although a performance gap persists relative to the strongest recent models, particularly Kumo-Tabular, TabPFN-3.5 \cite{tabpfn-3.5} and LimiX-2 \cite{limix-2}. In terms of computational cost, \textsc{OrionMoE-Small-v1.0} occupies an intermediate position among foundation models, without reaching the overall Pareto frontier. These results establish \textsc{OrionMoE-Small-v1.0} as a competitive compact alternative to several existing tabular foundation models, while leaving room for improvement in both predictive accuracy and inference efficiency.

\begin{table*}[t]
\centering
\small
\setlength{\tabcolsep}{4.5pt}
\caption{TabArena Elo scores across task and dataset regimes. Elo is recomputed independently within each subset. Higher is better. The number of datasets used to compute each score is shown in parentheses.}
\label{tab:tabarena_regime_elo}
\resizebox{\textwidth}{!}{%
\begin{tabular}{lcccccccc}
\toprule
& \multicolumn{1}{c}{Overall}
& \multicolumn{2}{c}{By Task}
& \multicolumn{2}{c}{By Dataset Size}
& \multicolumn{3}{c}{By Target} \\
\cmidrule(lr){2-2}
\cmidrule(lr){3-4}
\cmidrule(lr){5-6}
\cmidrule(lr){7-9}
Model
& Elo
& Binary 
& Multiclass
& Small
& Medium
& Balanced
& Imbalanced 
& Extreme \\

& ($N=38$)
& ($N=30$)
& ($N=8$)
& ($N=28$)
& ($N=10$)
& ($N=10$)
& ($N=16$)
&($N=12$) \\
\midrule
Kumo-Tabular & \textbf{1795} & \textbf{1777} & \textbf{1946} & \textbf{1776} & \textbf{2030} & 1743 & 1848 & \textbf{1905} \\
LimiX-2 & 1769 & 1761 & 1862 & 1764 & 1945 & \textbf{1794} & \textbf{1855} & 1747 \\
Kumo-Tabular-Medium & 1726 & 1719 & 1814 & 1731 & 1867 & 1699 & 1827 & 1729 \\
TabPFN-3.5 & 1677 & 1692 & 1672 & 1637 & 2009 & 1674 & 1750 & 1685 \\
TabFM & 1583 & 1646 & 1433 & 1604 & 1657 & 1601 & 1795 & 1446 \\
Kumo-Tabular-Small & 1558 & 1545 & 1655 & 1566 & 1670 & 1476 & 1657 & 1604 \\
TabPFN-3.5-Fast & 1554 & 1568 & 1544 & 1526 & 1809 & 1613 & 1597 & 1536 \\
Mitra-v2 & 1533 & 1525 & 1607 & 1526 & 1697 & 1467 & 1657 & 1532 \\
Causilo & 1530 & 1509 & 1668 & 1547 & 1612 & 1489 & 1620 & 1540 \\
EXAONE-Tabular & 1515 & 1509 & 1578 & 1501 & 1703 & 1419 & 1572 & 1627 \\
TabPFN-3 & 1351 & 1390 & 1254 & 1368 & 1405 & 1431 & 1451 & 1247 \\
OrionMoE-Small-v1.0 & 1320 & 1340 & 1276 & 1379 & 1253 & 1419 & 1470 & 1166 \\
XIAOMI-TabLDM & 1287 & 1296 & 1280 & 1327 & 1262 & 1283 & 1394 & 1225 \\
TabICL-v2 & 1277 & 1301 & 1221 & 1316 & 1253 & 1295 & 1368 & 1212 \\
RealTabPFN-V2.5 (T+E) & 1244 & 1215 & 1389 & 1297 & 1173 & 1291 & 1287 & 1197 \\
TabDPT-1.3 & 1215 & 1233 & 1169 & 1238 & 1217 & 1277 & 1337 & 1086 \\
TabM (T+E) & 1110 & 1109 & 1128 & 1103 & 1179 & 1128 & 1055 & 1184 \\
TabDPT-Turbo & 1090 & 1100 & 1066 & 1119 & 1035 & 1191 & 1172 & 942 \\
TabICL & 1072 & 1108 & 955 & 1081 & 1074 & 1182 & 1057 & 1021 \\
LightGBM (T+E) & 1071 & 1067 & 1102 & 1039 & 1214 & 1036 & 1069 & 1116 \\
LimiX & 1056 & 1018 & 1226 & 1149 & 783 & 1106 & 1083 & 993 \\
CatBoost (T+E) & 1048 & 1039 & 1094 & 994 & 1272 & 1007 & 1004 & 1151 \\
CatBoost & 1032 & 1050 & 972 & 970 & 1279 & 971 & 986 & 1153 \\
Mitra & 1019 & 1002 & 1094 & 1175 & 481 & 1088 & 1053 & 928 \\
LightXGB (T+E) & 1005 & 1000 & 1033 & 945 & 1231 & 945 & 966 & 1111 \\
TabSwift & 996 & 1022 & 903 & 1016 & 929 & 1038 & 1075 & 867 \\
TabM & 948 & 939 & 991 & 954 & 912 & 954 & 910 & 991 \\
SAP-RPT-OSS & 948 & 932 & 1018 & 994 & 768 & 994 & 873 & 1005 \\
TabDPT & 901 & 903 & 890 & 935 & 738 & 960 & 950 & 765 \\
RealMLP & 849 & 867 & 767 & 829 & 869 & 889 & 812 & 849 \\
XGBoost & 835 & 822 & 879 & 778 & 969 & 782 & 812 & 889 \\
RF (T+E) & 791 & 740 & 978 & 759 & 831 & 793 & 698 & 885 \\
LightGBM & 788 & 768 & 858 & 758 & 817 & 791 & 748 & 813 \\
iLTM & 766 & 785 & 663 & 727 & 818 & 719 & 759 & 786 \\
OrionMSP & 733 & 759 & 585 & 782 & 429 & 818 & 647 & 736 \\
TabSTAR & 647 & 684 & 406 & 679 & 409 & 694 & 700 & 421 \\
TabFlex & 626 & 655 & 436 & 663 & 356 & 672 & 601 & 562 \\
RF & 587 & 583 & 570 & 583 & 481 & 671 & 522 & 530 \\
\bottomrule
\end{tabular}%
}
\end{table*}

Table~\ref{tab:tabarena_regime_elo} provides a more detailed view of performance across dataset regimes. \textsc{OrionMoE-Small-v1.0} achieves an Elo rating of $1340$ on binary classification and $1276$ on multiclass classification, outperforming TabPFN-3 on the latter ($1254$). The distinction between dataset sizes is more pronounced. On small datasets, \textsc{OrionMoE-Small-v1.0} obtains an Elo of $1379$, exceeding TabPFN-3 ($1368$), TabICLv2 ($1316$), and Xiaomi-TabLDM ($1327$). Its relative position deteriorates on medium-sized datasets, where the strongest foundation models obtain substantially higher ratings.

Target-class imbalance reveals another informative pattern. \textsc{OrionMoE-Small-v1.0} reaches an Elo of $1470$ on imbalanced datasets, surpassing TabPFN-3 ($1451$), TabICLv2 ($1368$), and Xiaomi-TabLDM ($1394$). However, this advantage does not extend to the most extreme imbalance regime, where \textsc{OrionMoE-Small-v1.0} obtains a lower rating of $1166$. The results therefore indicate favorable relative performance under moderate class imbalance, rather than uniform robustness to increasingly skewed class distributions.

Overall, TabArena shows that \textsc{OrionMoE-Small-v1.0} achieves competitive performance against several established foundation models while remaining below the strongest models in absolute Elo. Its relative performance is
particularly encouraging on smaller classification problems and moderately imbalanced targets. These patterns motivate a broader evaluation of whether the same behavior persists across alternative benchmark distributions and more challenging generalization settings.


\subsection{BeyondArena}
\label{sec:beyondarena}

BeyondArena \cite{beyondarena} evaluates tabular learning methods beyond conventional small-scale IID settings, incorporating grouped and temporal distribution shifts, larger datasets, and varying feature dimensionalities. Grouped evaluation measures generalization to previously unseen subpopulations, while temporal evaluation tests predictions on observations collected after the training period. These settings provide a complementary perspective on the generalization capabilities of tabular foundation models, particularly when compared with conventional methods that can be optimized independently for each downstream task.

\begin{figure}[htbp]
    \centering
    \includegraphics[width=0.9\linewidth]{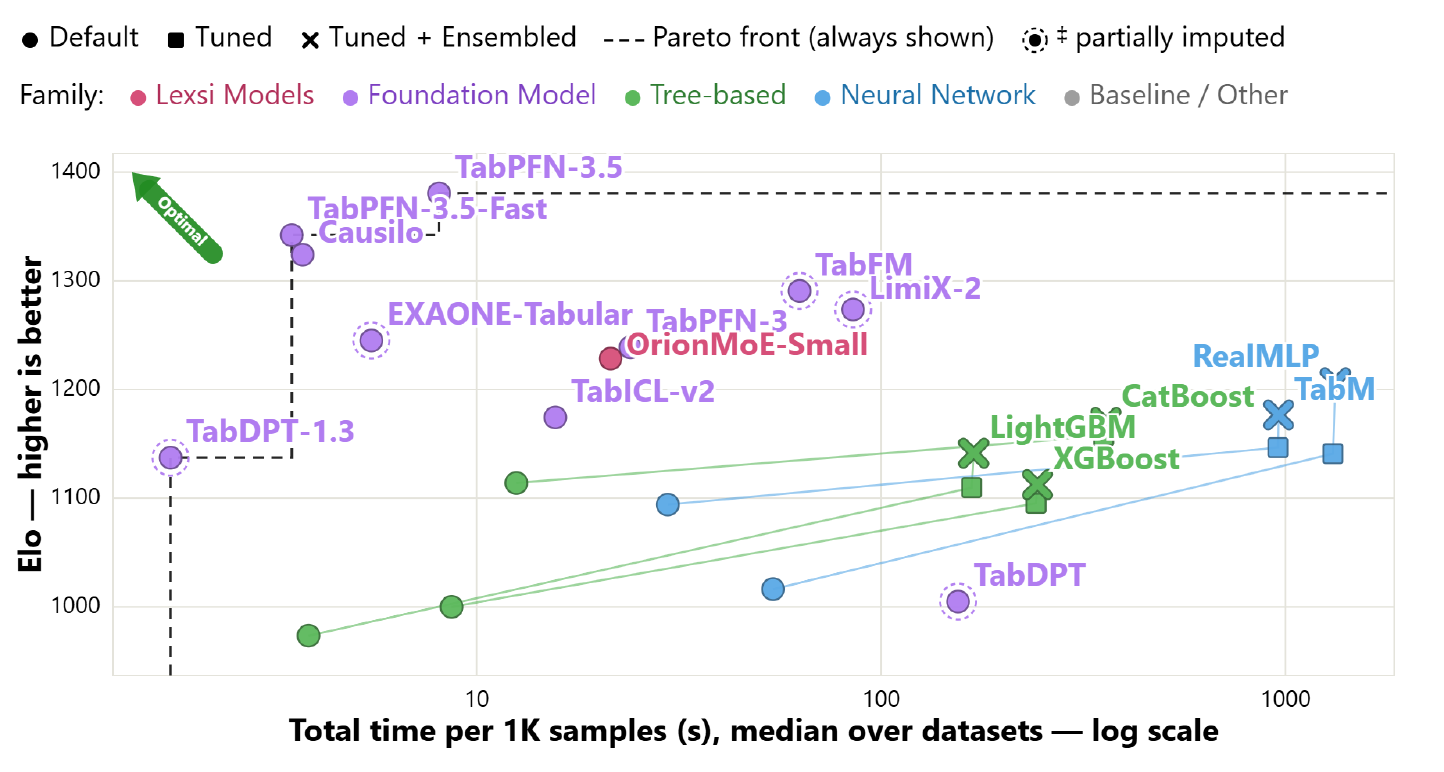}
    \caption{Pareto frontier of Elo rating versus total runtime per 1K samples on BeyondArena classification tasks. Higher Elo indicates stronger predictive performance and lower runtime indicates greater efficiency.  The frontier summarizes the trade-off between predictive quality and computational cost under more challenging IID, grouped, temporal, large-scale, and high-dimensional evaluation settings.}
    \label{fig:beyondarena-pareto}
\end{figure}

Figure~\ref{fig:beyondarena-pareto} presents the relationship between predictive performance and total runtime across BeyondArena classification tasks. \textsc{OrionMoE-Small-v1.0} obtains an overall Elo rating of $1149$ on the 98 evaluated datasets, ranking eighth among the compared methods. It outperforms several tabular foundation models, including TabICLv2 ($1089$), TabDPT-1.3 ($1041$), and RealTabPFN-V2.5 ($1048$), while approaching TabPFN-3 ($1166$). A larger gap remains relative to TabPFN-3.5 ($1348$) and Causilo \cite{causilo} ($1268$). \textsc{OrionMoE-Small-v1.0} also exceeds the overall Elo ratings of several tuned conventional baselines, including RealMLP, TabM, CatBoost, and LightGBM. 

\begin{figure}[htbp]
    \centering
    \includegraphics[width=1\linewidth]{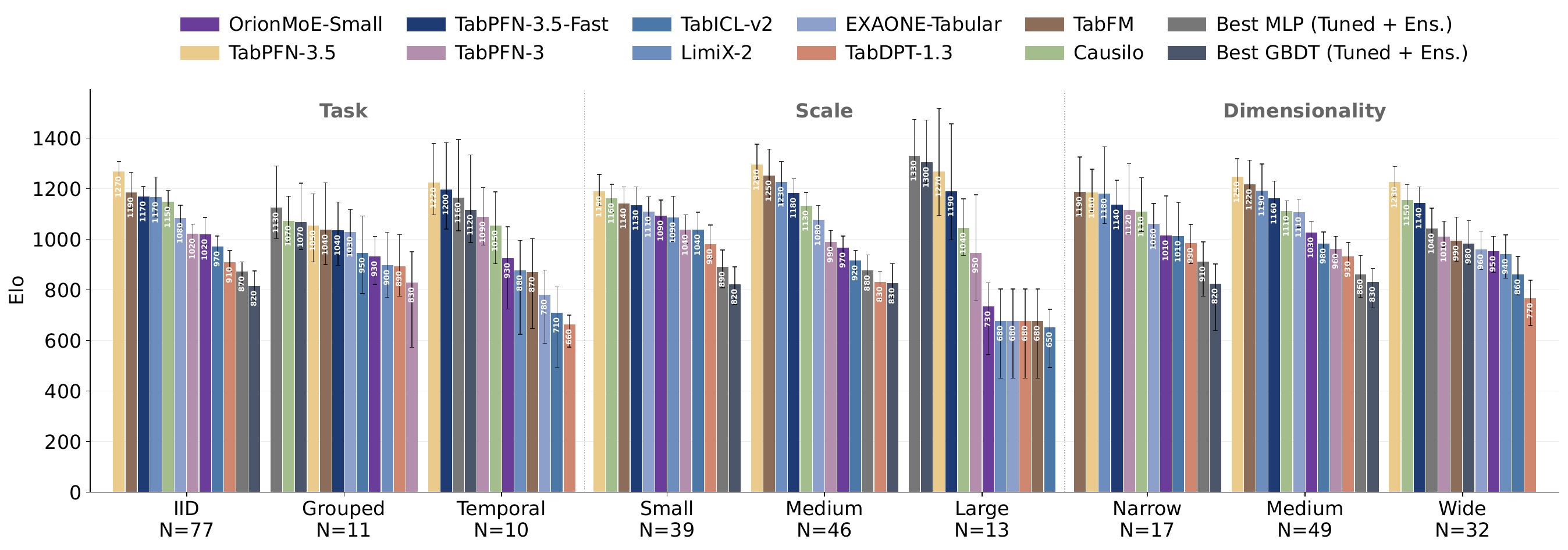}
    \caption{Elo ratings on BeyondArena across evaluation regime, dataset scale, and dimensionality.  Results are reported separately for IID, grouped, and temporal splits, as well as for different dataset-size and feature-dimensionality regimes. The decomposition highlights how the relative ranking of tabular models changes across distribution shifts and dataset characteristics.}
    \label{fig:beyondarena-ablation}
\end{figure}

The aggregate results conceal substantial differences between evaluation regimes. Figure~\ref{fig:beyondarena-ablation} and Table~\ref{tab:beyondarena_regime_elo} decompose performance by task type, distribution shift, dataset size, and dimensionality. On IID datasets, \textsc{OrionMoE-Small-v1.0} reaches an Elo rating of $1210$, outperforming several recent foundation models and conventional baselines, although remaining behind TabPFN-3.5 ($1444$), Causilo ($1342$), and TabPFN-3 ($1224$). Under grouped and temporal splits, its relative position becomes less favorable, with Elo ratings of $979$ and $1009$, respectively. In these regimes, tuned conventional methods become particularly competitive: RealMLP with tuning and ensembling achieves the highest Elo on both grouped ($1220$) and temporal ($1325$) evaluations. This contrasts with the IID regime, where foundation models occupy the leading positions, illustrating the challenge of transferring strong in-context prediction capabilities to shifted test distributions.

Dataset scale introduces a further distinction. \textsc{OrionMoE-Small-v1.0} obtains an Elo rating of $1225$ on small datasets, remaining competitive with several stronger overall models, including EXAONE-Tabular ($1242$) and Causilo ($1304$). On medium-sized datasets, it achieves $1206$, but on large datasets its Elo drops to $821$, substantially below the strongest tuned conventional systems. These results suggest that dataset scale is an important factor in the observed performance differences across BeyondArena regimes. The decomposition by feature dimensionality likewise shows that \textsc{OrionMoE-Small-v1.0} remains competitive with several established foundation models, although it does not consistently improve its relative ranking as dimensionality increases.

\begin{table*}[t]
\centering
\scriptsize
\setlength{\tabcolsep}{3.0pt}
\caption{BeyondArena Elo scores across task, split-regime, dataset-size, and dimensionality regimes. Elo is recomputed independently within each subset using datasets evaluated by all participating methods. Higher is better. $N$ is the number of datasets used to compute each score. T+E means Tuned + Ensemble.}
\label{tab:beyondarena_regime_elo}
\resizebox{\textwidth}{!}{%
\begin{tabular}{lcccccccccccc}
\toprule
& \multicolumn{1}{c}{Overall}
& \multicolumn{2}{c}{By Task}
& \multicolumn{3}{c}{By Regime}
& \multicolumn{3}{c}{By Dataset Size}
& \multicolumn{3}{c}{By Dimensionality} \\
\cmidrule(lr){2-2}
\cmidrule(lr){3-4}
\cmidrule(lr){5-7}
\cmidrule(lr){8-10}
\cmidrule(lr){11-13}
Model
& Elo 
& Binary 
& Multiclass
& IID 
& Grouped 
& Temporal
& Small
& Medium 
& Large
& Narrow 
& Medium
& Wide \\
& ($N=98$)
& ($N=73$)
& ($N=25$)
& ($N=77$)
& ($N=11$)
& ($N=10$)
& ($N=38$)
& ($N=47$)
& ($N=13$)
& ($N=13$)
& ($N=58$)
& ($N=27$) \\
\midrule
TabPFN-3.5 & \textbf{1348} & \textbf{1350} & \textbf{1349} & \textbf{1444} & 1095 & 1319 & \textbf{1326} & \textbf{1482} & 1294 & \textbf{1304} & \textbf{1404} & \textbf{1303} \\
TabPFN-3.5-Fast & 1287 & 1272 & 1343 & 1354 & 1083 & 1285 & 1275 & 1396 & 1219 & 1274 & 1333 & 1240 \\
Causilo & 1268 & 1275 & 1255 & 1342 & 1126 & 1123 & 1304 & 1356 & 1106 & 1241 & 1300 & 1248 \\
TabFM & 1222 & 1202 & 1295 & 1318 & 1076 & 924 & 1258 & 1412 & 727 & 1286 & 1302 & 1088 \\
LimiX-2 & 1188 & 1171 & 1246 & 1286 & 973 & 924 & 1195 & 1374 & 727 & 1269 & 1292 & 1008 \\
EXAONE-Tabular & 1176 & 1168 & 1207 & 1253 & 1085 & 855 & 1242 & 1290 & 727 & 1234 & 1236 & 1067 \\
TabPFN-3 & 1166 & 1168 & 1163 & 1224 & 899 & 1204 & 1185 & 1229 & 1017 & 1245 & 1185 & 1113 \\
\textsc{OrionMoE-Small-v1.0} & 1149 & 1148 & 1157 & 1210 & 979 & 1009 & 1225 & 1206 & 821 & 1128 & 1201 & 1079 \\
RealMLP (T+E) & 1117 & 1106 & 1153 & 1084 & \textbf{1220} & \textbf{1325} & 1061 & 1128 & \textbf{1449} & 1082 & 1098 & 1185 \\
TabICL-v2 & 1089 & 1075 & 1137 & 1148 & 966 & 822 & 1156 & 1133 & 748 & 1153 & 1159 & 940 \\
TabM (T=E) & 1076 & 1070 & 1098 & 1039 & 1198 & 1263 & 1012 & 1081 & 1429 & 967 & 1065 & 1161 \\
Catboost (T+E) & 1062 & 1069 & 1044 & 1037 & 1079 & 1270 & 996 & 1074 & 1365 & 1022 & 1060 & 1091 \\
Catboost (T) & 1049 & 1054 & 1034 & 1019 & 1070 & 1276 & 986 & 1053 & 1354 & 1031 & 1034 & 1090 \\
RealTabPFN-V2.5 & 1048 & 1031 & 1106 & 1086 & 979 & 833 & 1107 & 1081 & 727 & 1134 & 1055 & 1002 \\
LightGBM (T+E) & 1044 & 1053 & 1017 & 1005 & 1126 & 1260 & 951 & 1067 & 1393 & 983 & 1029 & 1106 \\
TabDPT-1.3 & 1041 & 1026 & 1089 & 1086 & 970 & 749 & 1122 & 1050 & 727 & 1201 & 1095 & 867 \\
TabM (T) & 1034 & 1027 & 1058 & 993 & 1184 & 1168 & 986 & 1020 & 1333 & 905 & 1021 & 1121 \\
RealMLP (T) & 1028 & 1016 & 1066 & 986 & 1141 & 1219 & 976 & 1021 & 1302 & 1031 & 1002 & 1080 \\
XGBoost (T+E) & 1003 & 1031 & 919 & 973 & 982 & 1266 & 853 & 1054 & 1406 & 846 & 991 & 1095 \\
LightGBM (T) & 1002 & 1013 & 968 & 966 & 1093 & 1139 & 923 & 1003 & 1322 & 965 & 975 & 1069 \\
CatBoost & 1001 & 1027 & 922 & 976 & 1009 & 1168 & 915 & 1020 & 1264 & 953 & 986 & 1050 \\
XGBoost (T) & 979 & 1006 & 895 & 946 & 975 & 1207 & 844 & 1008 & 1371 & 855 & 953 & 1080 \\
TabM & 978 & 976 & 985 & 941 & 1068 & 1117 & 975 & 940 & 1116 & 896 & 961 & 1044 \\
TabSwift & 935 & 954 & 877 & 953 & 885 & 764 & 984 & 918 & 727 & 967 & 952 & 875 \\
RealMLP & 877 & 894 & 819 & 829 & 914 & 1096 & 815 & 859 & 1024 & 878 & 835 & 941 \\
TabDPT & 876 & 880 & 863 & 886 & 861 & 678 & 967 & 781 & 736 & 1057 & 864 & 790 \\
XGBoost & 862 & 894 & 751 & 822 & 825 & 1104 & 672 & 904 & 1128 & 714 & 824 & 974 \\
LightGBM & 837 & 858 & 767 & 797 & 847 & 1011 & 739 & 819 & 1052 & 775 & 796 & 922 \\
RF (T+E) & 835 & 821 & 874 & 806 & 955 & 773 & 868 & 747 & 857 & 856 & 818 & 839 \\
RF & 761 & 756 & 772 & 713 & 960 & 688 & 843 & 613 & 727 & 727 & 753 & 766 \\
\bottomrule
\end{tabular}%
}
\end{table*}

\begin{figure}[htbp]
    \centering
    \includegraphics[width=1\linewidth]{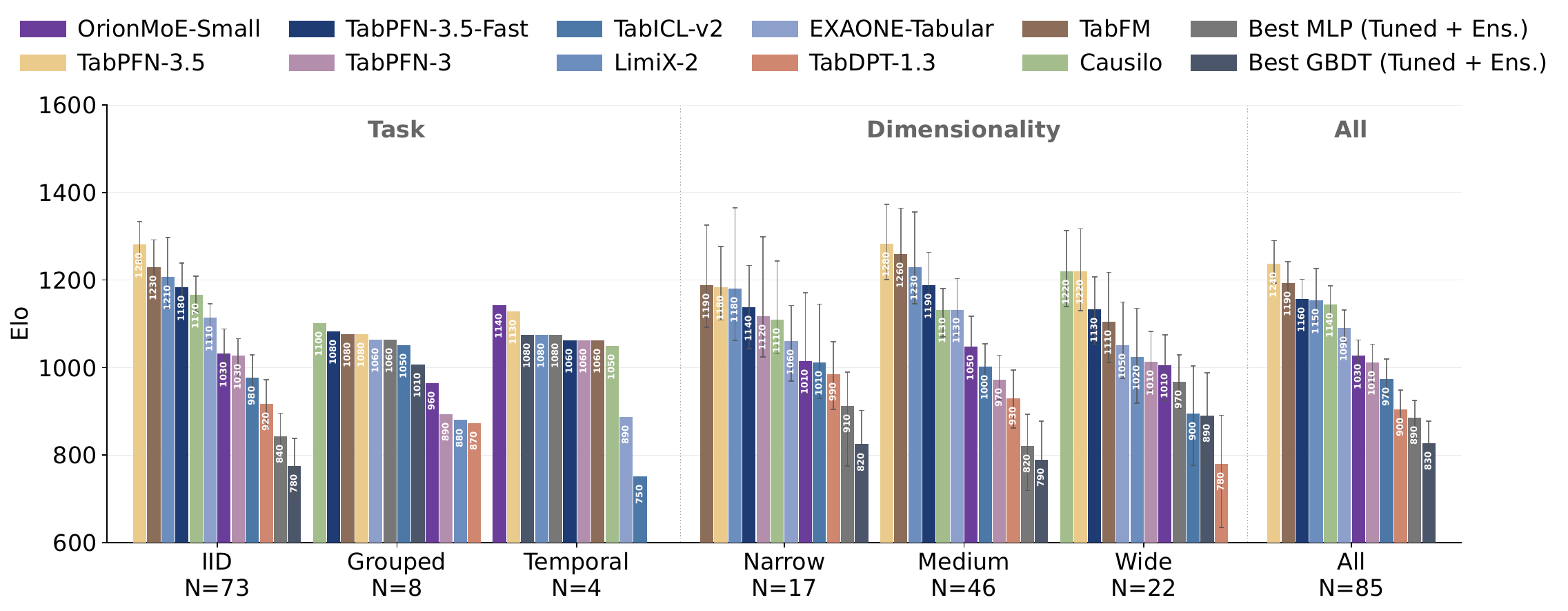}
    \caption{Elo ratings on BeyondArena after restricting the analysis to datasets with fewer than 100K samples. Results are decomposed by IID, grouped, and temporal evaluation regimes and by feature dimensionality.  Removing large-scale datasets isolates the effect of distribution shift and dimensionality from the additional difficulty introduced by dataset scale.}
    \label{fig:beyondarena-ablation-nolarge}
\end{figure}

Importantly, dataset scale and evaluation regime are not independent. To examine whether the apparent distribution-shift disadvantage is partly associated with the presence of very large datasets, we repeat the analysis after excluding datasets with more than $100{,}000$ samples. Figure~\ref{fig:beyondarena-ablation-nolarge} reports the resulting evaluation over 85 datasets. Restricting the analysis reduces the influence of large-scale problems and provides a complementary view of the relative rankings under IID, grouped, and temporal splits.

In this restricted setting, \textsc{OrionMoE-Small-v1.0} remains within the competitive foundation-model group on IID tasks. Its relative behavior on grouped and temporal tasks is more nuanced: the ranking varies across the
smaller subsets, and the reduced number of shifted datasets limits the strength of any general conclusion. The results indicate that the difficulties observed on the full BeyondArena suite cannot be attributed
exclusively to the presence of large datasets. Rather, performance depends jointly on the evaluation regime, dataset scale, and feature characteristics.

Taken together, BeyondArena highlights both the strengths and current limitations of OrionMoE-Small. The model maintains competitive predictive performance against several established foundation models on standard classification problems, but does not yet consistently match the strongest methods under grouped and temporal distribution shifts or on very large datasets. These findings also illustrate why aggregate benchmark rankings alone are insufficient to characterize the generalization behavior of tabular foundation models.


\subsection{OpenML-CC18}
\label{sec:openml_cc18}

OpenML-CC18~\citep{openmlcc18} is a widely used benchmark suite comprising 72 classification tasks spanning diverse dataset sizes, feature dimensionalities, and class distributions. We follow the standard evaluation protocol provided by the benchmark library, retaining its predefined train--test splits, cross-validation folds, and default evaluation settings without modification. This ensures consistency with the established OpenML-CC18 evaluation procedure. To enable a fair comparison across the broadest set of participating foundation models, we first report results on the 48 tasks successfully completed by all compared methods, ensuring that Elo ratings and model rankings are computed over identical benchmark instances. The complete list of the 48 datasets is provided in Appendix \ref{app:data-openml}.

Figure~\ref{fig:openml-elo} reports the corresponding Elo ratings. \textsc{OrionMoE-Small-v1.0} achieves an Elo of $1118$, behind Causilo ($1170$), and TabICLv2 ($1121$). In particular, \textsc{OrionMoE-Small-v1.0} is within $3$ Elo points of TabICLv2 and outperforms a broad set of recent foundation models, including TabLDM ($1083$), EXAONE-Tabular ($1066$), TabDPT-1.3 ($1061$), and TabPFN-3 ($1015$). These results place \textsc{OrionMoE-Small-v1.0} among the strongest small models on this common-task evaluation, demonstrating that its compact, dataset-conditioned architecture can achieve competitive predictive performance against recent dense tabular foundation models.

\begin{figure}[htbp]
    \centering
    \includegraphics[width=0.85\linewidth]{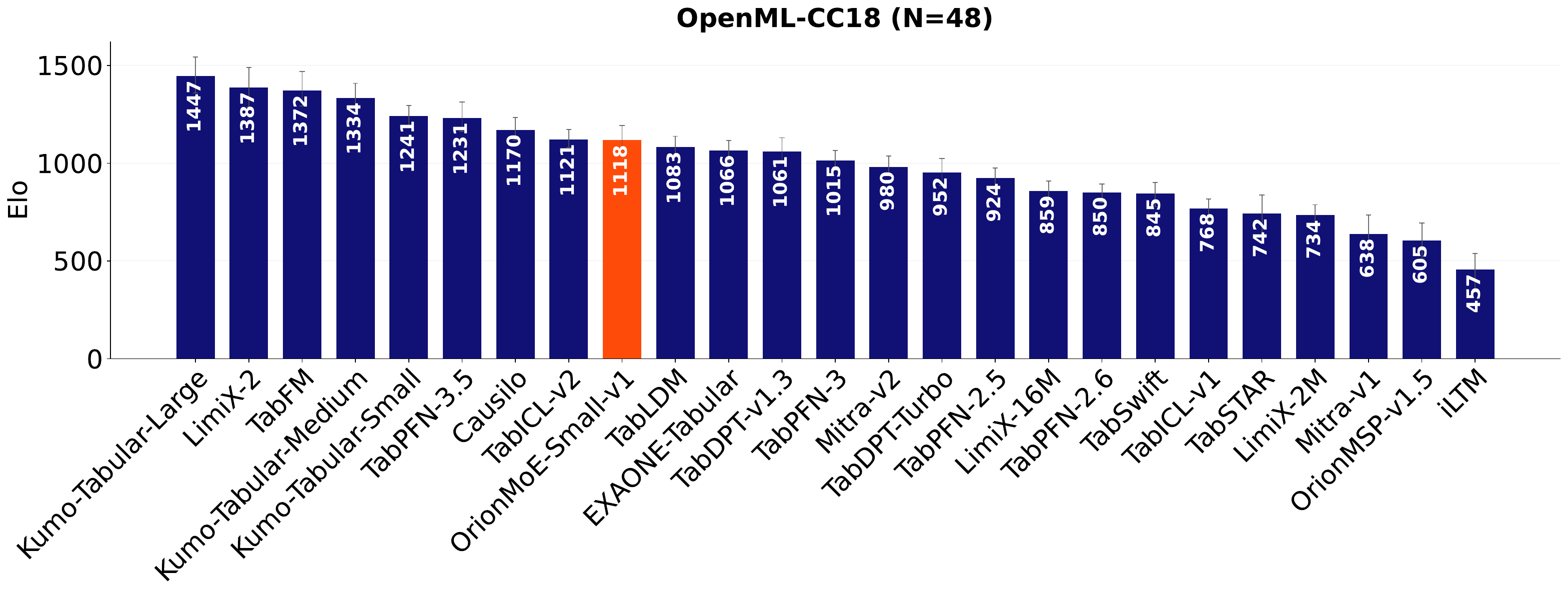}
    \caption{Elo ratings on the subset of OpenML-CC18 tasks completed by all compared models. Restricting the evaluation to common tasks ensures that pairwise comparisons and model rankings are computed over the same benchmark instances.}
    \label{fig:openml-elo}
\end{figure}


\subsection{TALENT}

TALENT~\citep{talent} provides a comprehensive collection of tabular learning datasets covering diverse classification problems, dataset scales, feature types, and statistical characteristics. We adopt the benchmark's original experimental protocol as implemented in the TALENT library, using its predefined training, validation, and test partitions and default evaluation settings. No additional dataset resplitting is performed. For the common-task comparison, we retain the 148 classification datasets successfully evaluated by all participating models, ensuring consistent Elo computation and direct comparisons across methods. We provide in Appendix \ref{app:data-talent} the list of these datasets.

\begin{figure*}[htbp]
    \centering
    \begin{subfigure}[t]{0.49\textwidth}
        \centering
        \includegraphics[width=1\linewidth]{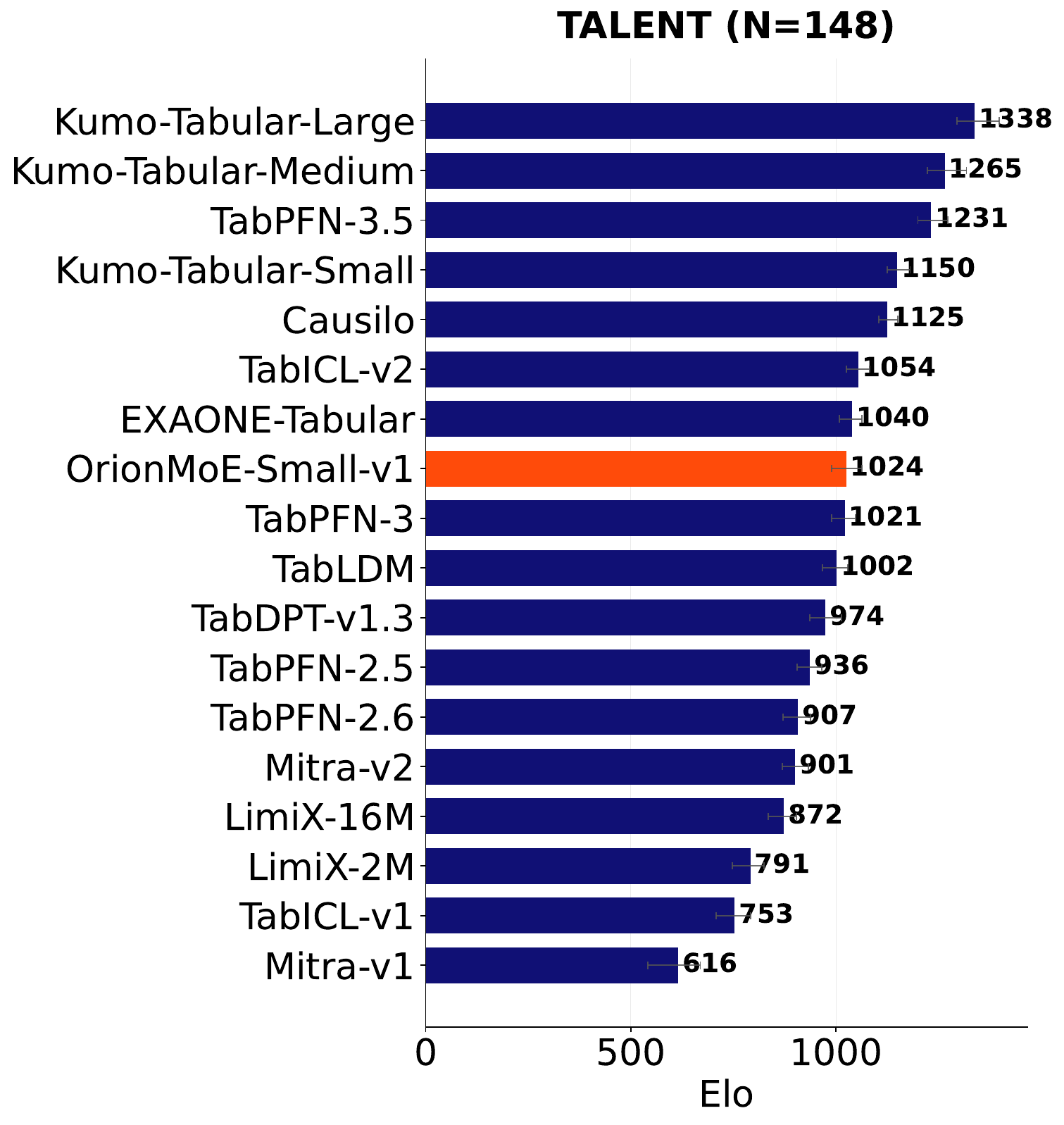}
        \caption{Elo ratings on common TALENT tasks.}
        \label{fig:talent-elo}
    \end{subfigure}
    \hfill
    \begin{subfigure}[t]{0.49\textwidth}
        \centering
        \includegraphics[width=1\linewidth]{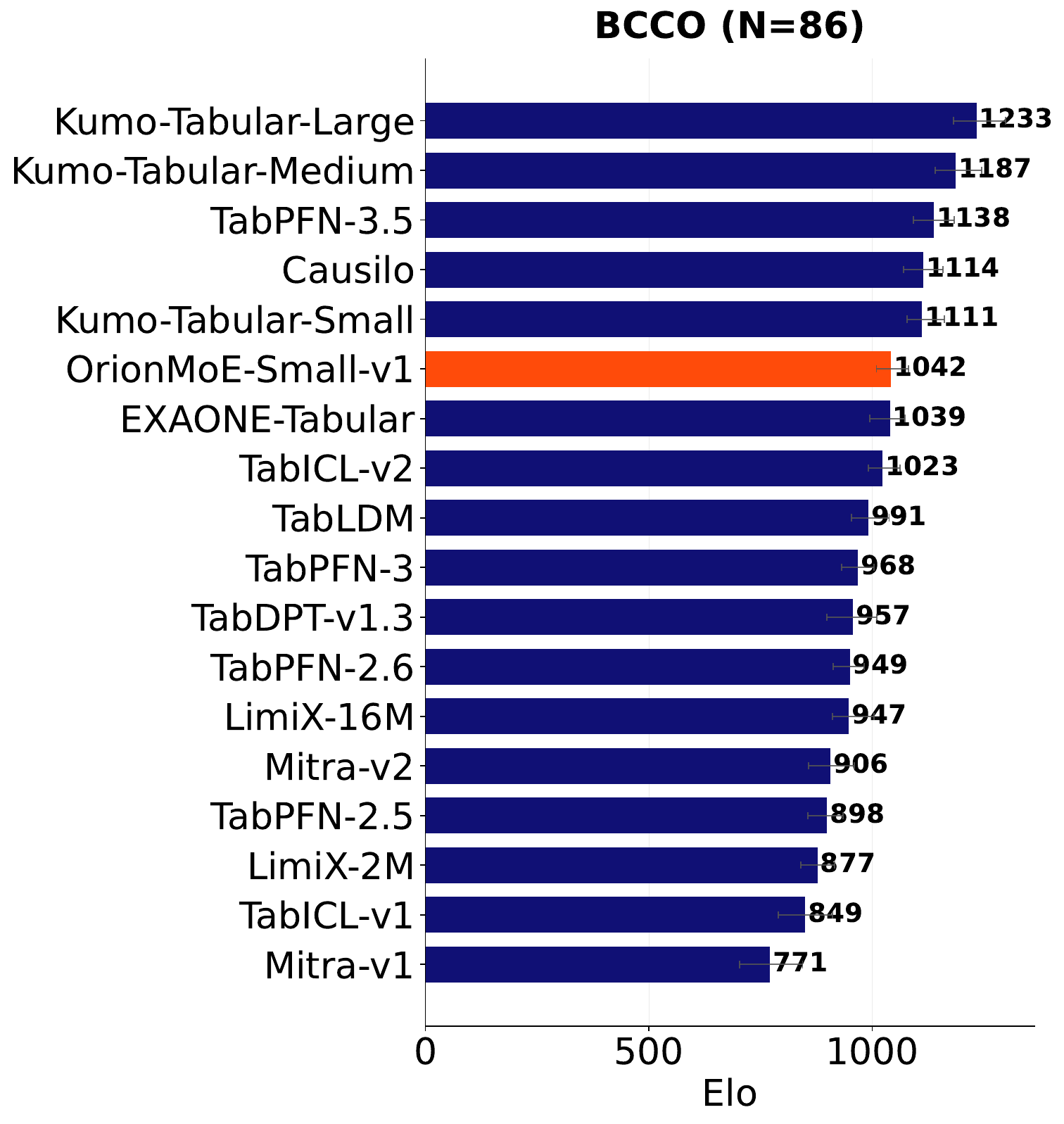}
        \caption{Elo ratings on common BCCO tasks.}
        \label{fig:bcco-elo}
    \end{subfigure}
    \caption{Elo ratings on the common-task subsets of TALENT and BCCO. \textbf{(a)} TALENT and \textbf{(b)} BCCO. For each benchmark, only tasks completed by all compared models are retained, ensuring that the reported Elo rankings are based on a consistent set of evaluation tasks.}
    \label{fig:talent-bcco-elo}
\end{figure*}

As shown in Figure~\ref{fig:talent-elo}, \textsc{OrionMoE-Small-v1.0} obtains an Elo rating of $1024$. It remains close to EXAONE-Tabular ($1040$) and TabICLv2 ($1054$), while outperforming TabPFN-3 ($1021$), TabLDM ($1002$), TabDPT-v1.3 ($974$), and several other recent foundation models. The strongest ratings are achieved by Kumo-Tabular models Large and Medium ($1338$) and ($1265$) respectively, indicating that a performance gap remains relative to the leading methods. Nevertheless, \textsc{OrionMoE-Small}'s position across this considerably broader dataset collection reinforces its competitiveness beyond the more constrained OpenML-CC18 evaluation.

\subsection{BCCO}

BCCO~\citep{limix} provides a complementary evaluation suite of tabular classification problems spanning heterogeneous dataset characteristics and predictive settings. We follow the default experimental protocol of the BCCO benchmark, preserving its predefined data partitions, evaluation folds where applicable, and standard evaluation settings. This avoids introducing additional variation from custom data splitting or protocol modifications. We initially evaluate the 86 datasets shared by all participating models as described in Appendix \ref{app:data-bcco}, allowing Elo ratings and model rankings to be computed over a common set of benchmark instances.

Figure~\ref{fig:bcco-elo} shows that \textsc{OrionMoE-Small-v1.0} obtains an Elo rating of $1042$, ranking sixth among the evaluated models. Its performance is particularly close to Kumo-Tabular-Small ($1111$), while exceeding EXAONE-Tabular ($1039$), TabICLv2 ($1023$), TabLDM ($991$), TabPFN-3 ($968$), and TabDPT-v1.3 ($957$). The leading methods remain Kumo-Tabular-Large ($1233$), Kumo-Tabular-Medium ($1187$) and TabPFN-3.5 ($1138$) . Importantly, \textsc{OrionMoE-Small-v1.0} maintains a strong relative position on this additional benchmark, suggesting that its competitive performance is not confined to a single benchmark collection.

Taken together, the common-task evaluations demonstrate that \textsc{OrionMoE-Small-v1.0} consistently ranks among the stronger recent tabular foundation models. While it does not surpass the leading models, it remains competitive with TabICLv2, EXAONE-Tabular, and other recent alternatives across three independently constructed benchmark suites.

\subsection{Evaluation on the Full Benchmark Suites}
\label{sec:full_benchmarks}

The common-task evaluations allow comparisons against a broad collection of tabular foundation models, but necessarily exclude datasets that some methods cannot successfully process. To complement these results, we conduct a full-benchmark evaluation restricted to models with complete coverage of each benchmark suite. This enables assessment over all 72 OpenML-CC18 tasks, 181 TALENT datasets, and 106 BCCO datasets, respectively. The two evaluation settings are complementary: common-task comparisons maximize model coverage, while full-benchmark comparisons maximize dataset coverage.

\begin{figure}[htbp]
    \centering
    \begin{subfigure}[]{0.49\textwidth}
        \centering
        \includegraphics[width=\linewidth]{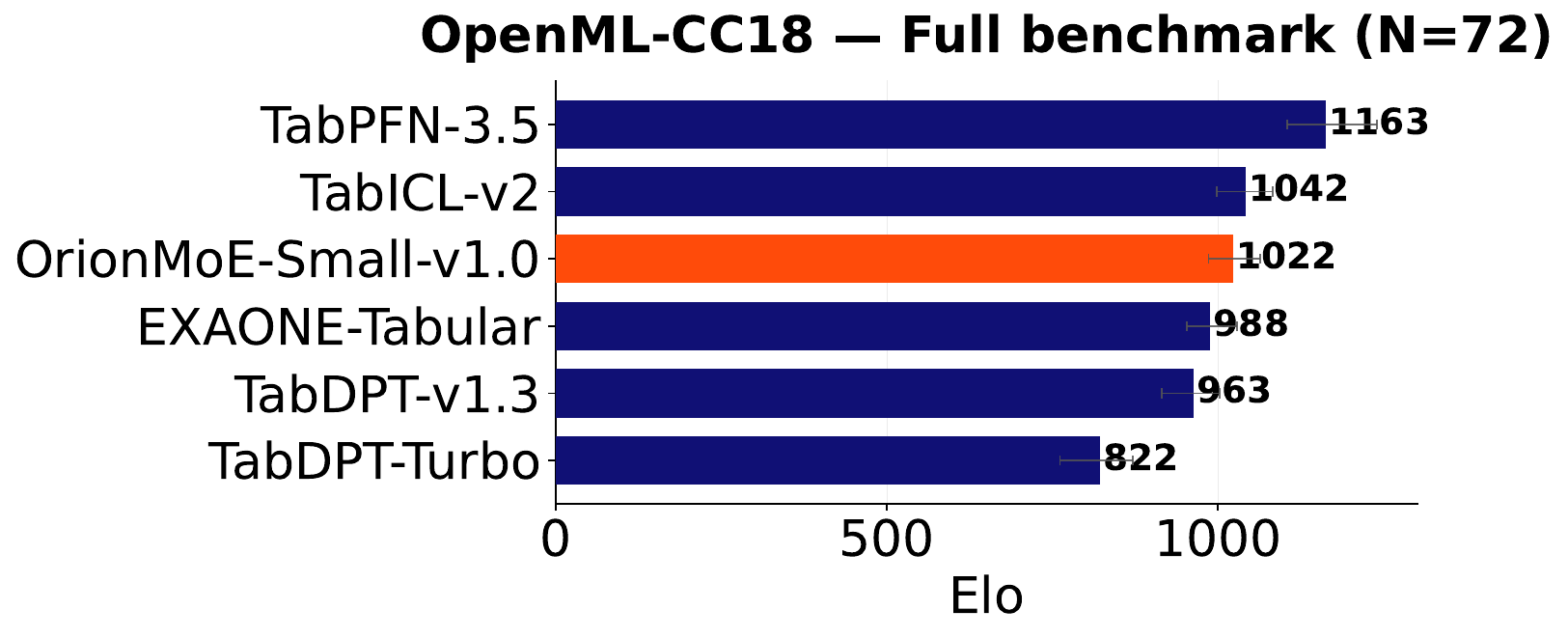}
        \caption{OpenML-CC18: models completing the full benchmark (72 datasets).}
        \label{fig:openml_full}
    \end{subfigure}
    \hfill
    \begin{subfigure}[]{0.49\textwidth}
        \centering
        \includegraphics[width=\linewidth]{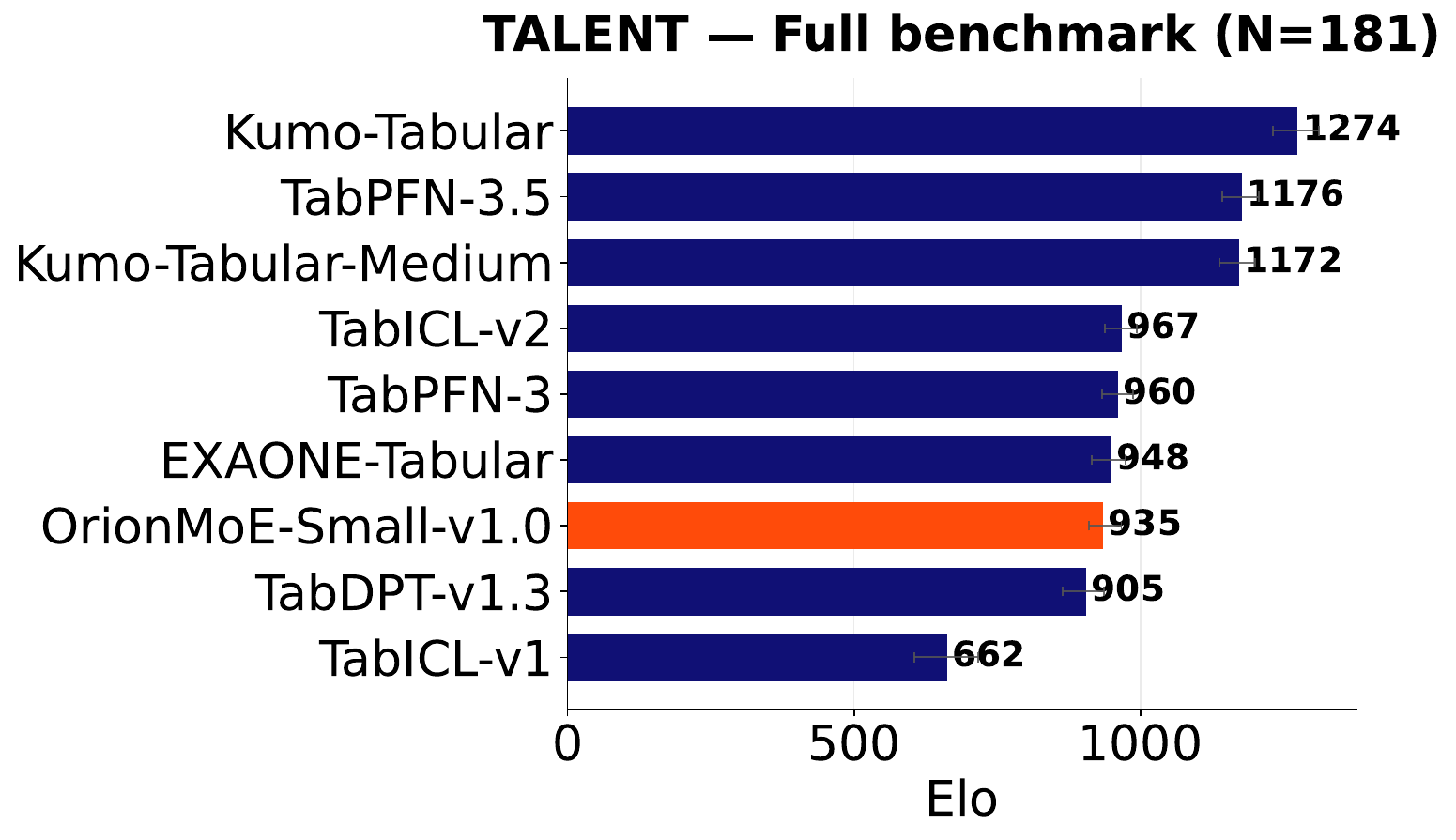}
        \caption{TALENT: models completing the full benchmark (181 datasets).}
        \label{fig:talent_full}
    \end{subfigure}
    \caption{ Elo ratings for models that completed the full benchmark suite. \textbf{(a)} OpenML-CC18 and \textbf{(b)} TALENT. By restricting the comparison to models with complete benchmark coverage, the resulting rankings avoid differences caused by missing evaluations or partial task completion.}
    \label{fig:full_benchmarks}
\end{figure}

\begin{figure}[htbp]
    \centering
    \includegraphics[width=0.80\linewidth]{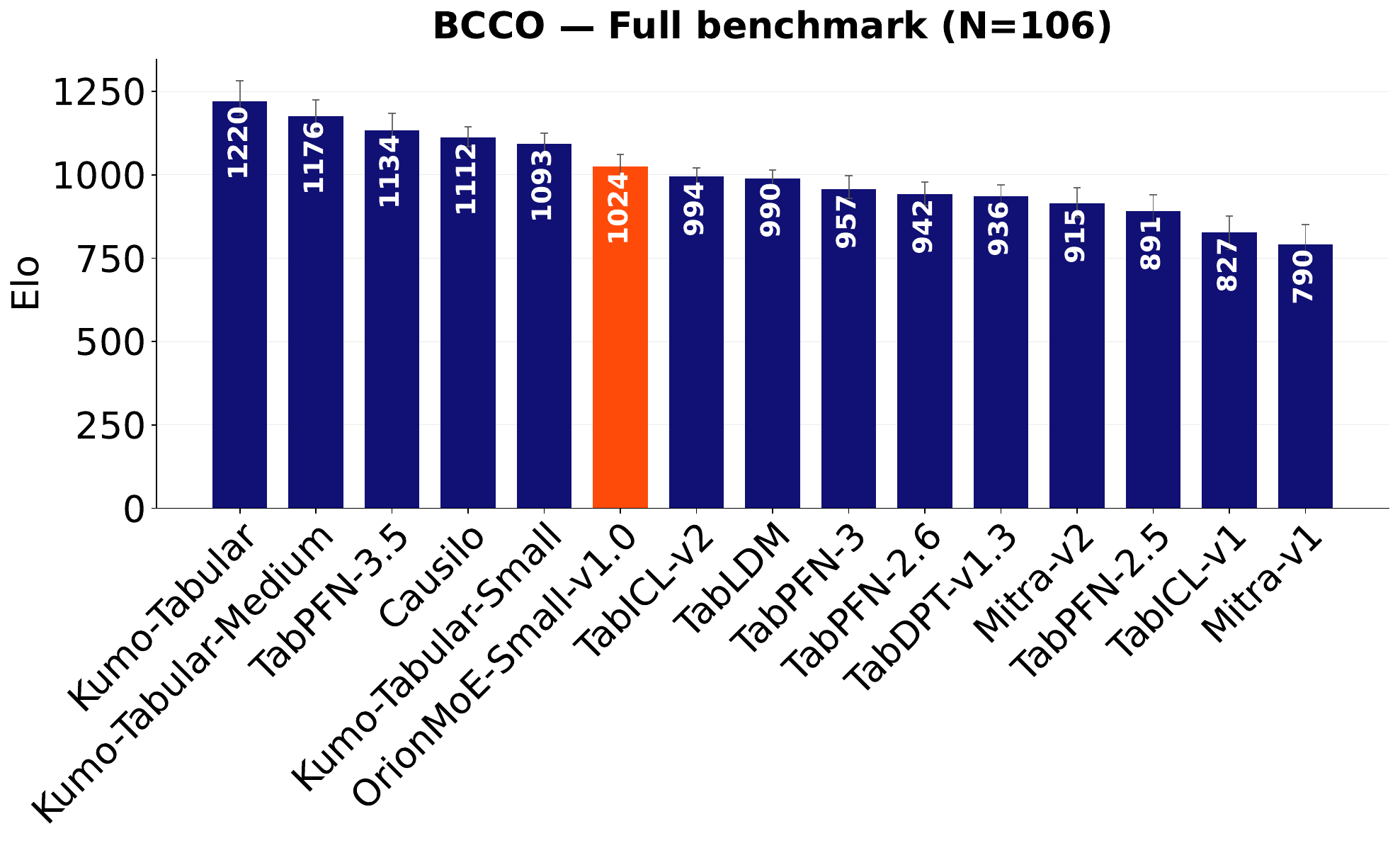}
    \caption{Elo ratings on BCCO for models that completed the full benchmark suite (106 datasets). Restricting the comparison to fully evaluated models provides a direct ranking without relying on partially observed benchmark results.}
    \label{fig:bcco_full}
\end{figure}

Figures~\ref{fig:bcco_full} and \ref{fig:full_benchmarks} present the Elo ratings of models completing the full evaluation suites. On OpenML-CC18 (Figure \ref{fig:openml_full}), \textsc{OrionMoE-Small-v1.0} achieves an Elo rating of $1022$, ranking third behind TabPFN-3.5 ($1163$) and very close to TabICLv2 ($1042$). It outperforms EXAONE-Tabular ($988$), TabDPT-v1.3 ($963$), and TabDPT-Turbo ($822$). Notably, its position remains competitive when the evaluation is expanded from the common 48-task subset to all 72 tasks.

On TALENT, \textsc{OrionMoE-Small-v1.0} obtains an Elo rating of $935$ across all 181 datasets. It remains close to EXAONE-Tabular ($948$), while exceeding TabDPT-v1.3 ($905$) and TabICLv1 ($662$). The leading foundation models, particularly large models (e.g., Kumo-Tabular models and TabPFN-3.5), retain an advantage, indicating that the broader dataset coverage does not eliminate the performance gap observed in the common-task comparison.

The full BCCO evaluation provides another encouraging result. Across all 106 datasets, \textsc{OrionMoE-Small-v1.0} achieves an Elo rating of $1024$, ranking six among the evaluated models. It trails TabPFN-3.5 ($1134$) and Causilo ($1112$), while outperforming TabICLv2 ($994$), TabLDM ($990$), TabPFN-3 ($957$), and the remaining baselines. This result is consistent with the strong relative position observed on the common BCCO subset.

Overall, these results show that \textsc{OrionMoE-Small-v1.0} maintains competitive predictive performance when evaluation is extended to complete benchmark collections. Although the participating models differ across the common-task and full-benchmark settings, the full-suite results demonstrate that \textsc{OrionMoE-Small-v1.0} can operate across the complete range of datasets considered in these three benchmarks while remaining competitive with recent tabular foundation models.

\subsection{Performance Across Model Capacity Regimes}
\label{sec:model_capacity}

To better understand the relationship between model capacity and predictive performance, we stratify the evaluated tabular foundation models into three parameter-count regimes: small ($<50$M), medium ($50$--$150$M), and large ($>150$M). This analysis complements the aggregate benchmark comparisons by distinguishing performance against similarly sized architectures from performance against models with substantially larger parameter budgets. We use parameter count as a measure of model capacity, while recognizing that it does not directly capture active computation, pretraining cost, or inference efficiency.

\begin{table}[t]
\centering
\small
\setlength{\tabcolsep}{5pt}
\caption{ Comparison of small tabular foundation models ($<50$M parameters) across four classification
benchmarks. Elo ratings are computed independently for each benchmark; higher is better. The strongest result within each benchmark is highlighted in bold.}
\label{tab:small_model_comparison}
\begin{tabular}{lccccc}
\toprule
\textbf{Model}
& \textbf{Params (M)}
& \textbf{TabArena}
& \textbf{BeyondArena}
& \textbf{OpenML-CC18}
& \textbf{TALENT} \\
\midrule
Causilo & 36.1 & 1530 & \textbf{1268} & 1170 & 1125 \\

\rowcolor{gray!12}
\textbf{OrionMoE-Small-v1.0} & \textbf{37.7} & 1320 & 1149 & 1118 & 1024 \\

TabICLv2 & 27.6 & 1277 & 1089 & 1121 & 1054 \\

TabICLv1 & 25 & 1072 & -- & 768 & 753 \\

Limix-16M & 16 & 1056 & -- & 859 & 872 \\

EXAONE-Tabular & 20.8 & 1515 & 1176 & 1066 & 1040 \\

Kumo-Tabular-Small & 28 & \textbf{1558} & -- & \textbf{1241} & \textbf{1150} \\

\bottomrule
\end{tabular}%
\end{table}

\paragraph{Comparison with small models.}
Table~\ref{tab:small_model_comparison} compares \textsc{OrionMoE-Small-v1.0} with foundation models containing fewer than 50 million parameters. With $37.7$M parameters, \textsc{OrionMoE-Small} achieves Elo ratings of $1320$ on TabArena, $1149$ on BeyondArena, $1118$ on OpenML-CC18, and $1024$ on TALENT. Among the small models, Kumo-Tabular-Small ($28$M) consistently achieves the strongest Elo ratings, indicating that substantial performance differences remain even at comparable model capacities.

Relative to TabICLv2 ($28$M), \textsc{OrionMoE-Small-v1.0} obtains higher ratings on TabArena and BeyondArena, while TabICLv2 retains a modest advantage on OpenML-CC18 and TALENT. \textsc{OrionMoE-Small} also outperforms the smaller EXAONE-Tabular ($20.8$M) on OpenML-CC18, although EXAONE achieves higher ratings on the other benchmarks. Compared with earlier small-scale models, including TabICLv1 and LimiX-16M, our model achieves higher ratings across all benchmarks for which results are available. Overall, these comparisons place \textsc{OrionMoE-Small-v1.0} among the competitive compact tabular foundation models, while highlighting the importance of architectural and training differences beyond parameter count alone.

\paragraph{Comparison with medium-sized models.}
Table~\ref{tab:medium_large_elo} positions \textsc{OrionMoE-Small-v1.0} against models containing between 50 and 150 million parameters. Despite having approximately 29\% fewer parameters than TabPFN-3 ($53$M), \textsc{OrionMoE-Small} remains close on TabArena ($1320$ versus $1351$) and BeyondArena ($1149$ versus $1166$), while achieving higher Elo ratings on OpenML-CC18 ($1118$ versus $1015$) and TALENT ($1024$ versus $1021$).

\textsc{OrionMoE-Small-v1.0} also outperforms TabDPT-1.3 ($63$M) across all four benchmarks, despite containing approximately 40\% fewer parameters. A similar pattern emerges against several other medium-sized foundation models: \textsc{OrionMoE-Small-v1.0} outperforms Xiaomi-TabLDM ($70$M) on TabArena, OpenML-CC18, and TALENT, and obtains higher ratings than Mitra-v1 and Mitra-v2 on the available OpenML-CC18 and TALENT evaluations. However, Mitra-v2 achieves a substantially higher rating on TabArena ($1533$ versus $1320$), demonstrating that the relative advantages are benchmark-dependent.

These results indicate that \textsc{OrionMoE-Small-v1.0} can achieve competitive performance against models with substantially larger parameter budgets, although the magnitude and direction of the differences depend on the evaluation distribution.

\begin{table*}[t]
\centering
\small
\setlength{\tabcolsep}{8pt}
\caption{
Comparison of \textsc{OrionMoE-Small} with medium-sized
($50$--$150$M parameters) and large
($>150$M parameters) tabular foundation models.
Elo ratings are reported on TabArena ($N=38$),
BeyondArena ($N=98$), OpenML-CC18 ($N=48$),
and TALENT ($N=148$). Higher is better.
Ratings are computed independently within each
benchmark and are not directly comparable across
benchmarks. A dash indicates unavailable results.
}
\label{tab:medium_large_elo}
\begin{tabular}{lccccc}
\toprule
\textbf{Model}
& \textbf{Params (M)}
& \textbf{TabArena}
& \textbf{BeyondArena}
& \textbf{OpenML-CC18}
& \textbf{TALENT} \\
\midrule

\rowcolor{gray!12}
\textbf{\textsc{OrionMoE-Small-v1.0}}
& \textbf{37.7} & 1320 & 1149 & 1118 & 1024 \\

\midrule
\multicolumn{6}{l}{
\textit{Medium models ($50$--$150$M)}
} \\

TabPFN-3
& 53 & 1351 & \textbf{1166} & 1015 & 1021 \\

TabDPT-1.3
& 63 & 1215 & 1041 & 1061 & 974 \\

TabDPT-Turbo
& 63 & 1090 & -- & 952 & -- \\

Mitra-v1
& 63 & 1019 & -- & 638 & 616 \\

Mitra-v2
& 63 & 1533 & -- & 980 & 901 \\

Xiaomi-TabLDM
& 70 & 1287 & -- & 1083 & 1002 \\

Kumo-Tabular-Medium & 62 & \textbf{1726} & -- & \textbf{1334} & \textbf{1265} \\

\midrule
\multicolumn{6}{l}{
\textit{Large models ($>150$M)}
} \\

TabPFN-3.5
& 220 & 1677 & \textbf{1348} & 1231 & 1231 \\

LimiX-2
& 400 & 1769 & 1188 & 1387 & -- \\

TabFM
& 408.7 & 1583 & 1222 & 1372 & -- \\

Kumo-Tabular-Large & 215 & \textbf{1795} & -- & \textbf{1447} & \textbf{1338} \\

\bottomrule
\end{tabular}
\end{table*}

\paragraph{Comparison with large models.}
The largest models in our comparison include Kumo-Tabular-Large ($215$M), TabPFN-3.5 ($220$M), LimiX-2 ($400$M), and TabFM ($408.7$M), representing approximately $5.7$--$10.8\times$ the parameter count of \textsc{OrionMoE-Small-v1.0}. These models generally achieve stronger predictive performance, particularly on TabArena and BeyondArena. TabPFN-3.5 obtains the highest Elo ratings on BeyondArena, OpenML-CC18, and TALENT, while LimiX-2 leads the TabArena comparison.

Nevertheless, the relationship between parameter count and performance is not monotonic. For example, Kumo-Tabular-Medium achieves higher Elo ratings than the larger TabPFN-3.5, LimiX-2 and TabFM models on TabArena, despite having substantially fewer parameters. Thus, increased capacity alone does not fully explain the relative performance of recent tabular foundation models.

\paragraph{Discussion.}
Overall, the capacity-stratified evaluation shows that \textsc{OrionMoE-Small-v1.0} remains competitive within the small-model category and can match or outperform several medium-sized foundation models across multiple benchmarks. At the same time, Causilo's stronger results at a comparable parameter count, together with the persistent advantage of the leading large models, indicate that substantial opportunities for improvement remain. The results support the practical viability of compact, dataset-conditioned sparse architectures, but do not establish that sparse routing itself is responsible for the observed performance differences. A controlled comparison with parameter-matched dense architectures would be required to isolate that contribution.

\section{License and Availability}
\label{sec:license}

\textsc{OrionMoE-Small-v1.0} is made available through a hosted inference API provided by Lexsi Labs. Access to and use of the API are governed by the \textit{Lexsi Labs Source Available License (LSAL) v1.2, OrionMoE Edition}. The model weights and underlying implementation are not publicly distributed, and inference is performed through the hosted service.

\paragraph{API access.}
The API enables users to perform in-context classification by providing labeled support observations and unlabeled query observations, without task-specific parameter updates. Access is subject to compliance with the OrionMoE license. The applicable license terms govern the permitted use of the API and its outputs.

\paragraph{Research use.}
The license provides for personal noncommercial research, academic research, teaching, and scholarly publication, subject to its applicable conditions and restrictions. Researchers may use the API to evaluate \textsc{OrionMoE-Small-v1.0}, conduct experiments, and report results in scientific publications, provided their activities comply with the license.

\paragraph{Organizational and commercial use.}
Organizational activities, including internal evaluation, benchmarking, and use in developing or evaluating an organization's models, products, or services, are subject to Section~1A of the OrionMoE license. Such activities require prior written permission or submission of a written acknowledgement identifying the organization and intended use, together with an attribution undertaking, which Lexsi Labs may accept in place of negotiated permission. These provisions do not independently authorize commercial exploitation. Commercial use requires a separate commercial license, while production or user-facing deployment requires express written authorization.

\paragraph{Code and model weights.}
The pretrained model weights and training implementation are not publicly distributed as part of this release. The architectural and methodological descriptions provided in this report are intended to support scientific understanding, evaluation, and further research into dataset-conditioned sparse computation.

\paragraph{License and enquiries.}
The complete LSAL v1.2, OrionMoE Edition license governs API usage and should be consulted for the definitive terms and conditions. For API access, organizational acknowledgements, research collaborations, or commercial licensing enquiries, please contact \href{mailto:support@lexsi.ai}{support@lexsi.ai}.

\section{Conclusion}
\label{sec:conclusion}

We introduced \textsc{OrionMoE-Small-v1.0}, a compact tabular foundation model that combines in-context learning with dataset-conditioned sparse mixture-of-experts computation. Built on a dual-axis transformer, \textsc{OrionMoE-Small} introduces conditional expert routing along enabling the model to adapt not only its representations but also the computational pathways applied to different features and datasets.

Across TabArena, BeyondArena, OpenML-CC18, TALENT, and BCCO, \textsc{OrionMoE-Small-v1.0} achieves competitive predictive performance against a diverse range of tabular foundation models without task-specific parameter updates. Our capacity-stratified evaluation further shows that, with only 37.7 million parameters, \textsc{OrionMoE-Small} remains competitive with several medium-sized models and outperforms larger alternatives on selected benchmarks. Nevertheless, the strongest models retain an advantage in overall predictive performance. 

We make \textsc{OrionMoE-Small-v1.0} available through the Lexsi inference API to facilitate research evaluations and practical exploration of dataset-conditioned sparse computation, while continuing to investigate more scalable and adaptive architectures for tabular in-context learning.

\newpage
\bibliographystyle{unsrt}
\bibliography{references}

\newpage
\appendix

\section{Additional Results}
\label{app:results}

\subsection{TabArena}
\label{app:tabarena}

\begin{figure*}[htbp]
    \centering
    \begin{subfigure}[t]{0.49\textwidth}
        \centering
        \includegraphics[width=1.07\linewidth]{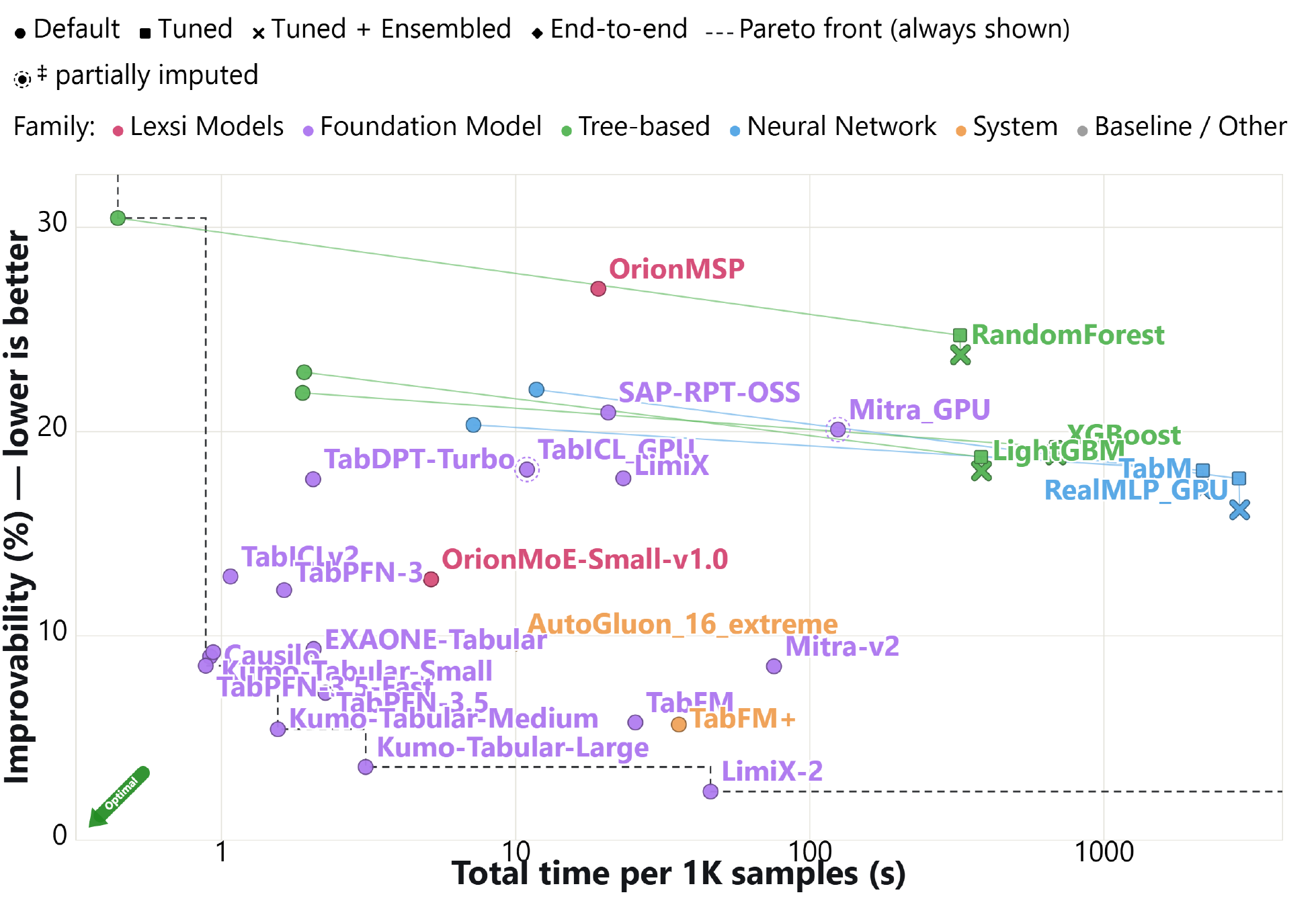}
        \caption{Improvability versus total runtime per 1K samples.}
        \label{fig:tabarena-improv-total}
    \end{subfigure}
    \hfill
    \begin{subfigure}[t]{0.49\textwidth}
        \centering
        \includegraphics[width=1.1\linewidth]{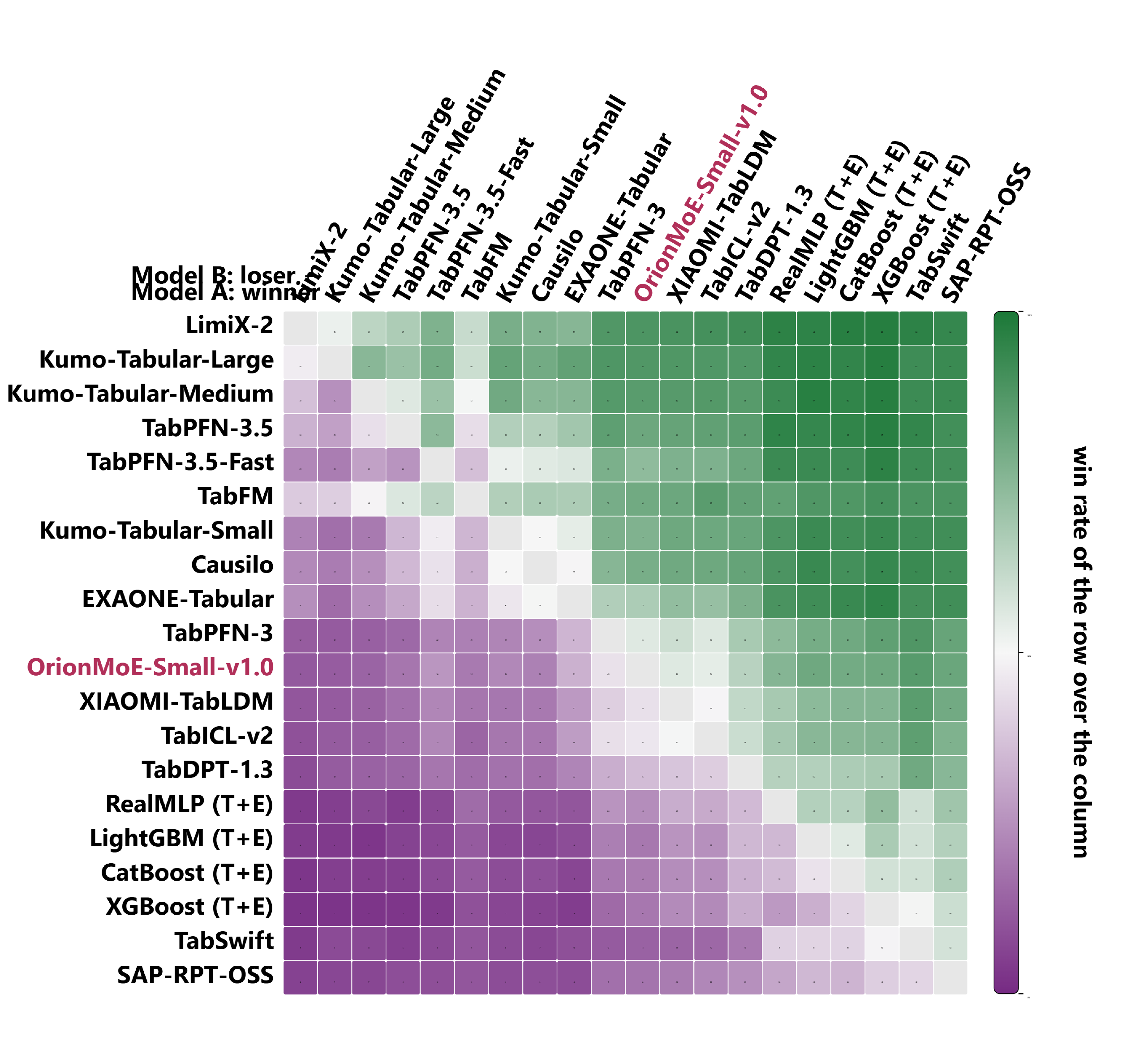}
        \caption{Pairwise win-rate matrix.}
        \label{fig:tabarena-winrate-matrix}
    \end{subfigure}
    \caption{Complementary performance analysis on TabArena classifiCation tasks. \textbf{(a)} Improvability as a function of total runtime per 1K samples, illustrating the trade-off between remaining performance headroom and computational cost. \textbf{(b)} Pairwise win-rate matrix showing how frequently each method outperforms the others across the evaluated tasks.}
    \label{fig:tabarena-efficiency}
\end{figure*}

We complement the main TabArena \cite{tabarena} evaluation with an analysis of improvability, pairwise win rates, and computational cost. Figure~\ref{fig:tabarena-efficiency} presents improvability as a function of total runtime and the pairwise win-rate matrix, while Table~\ref{tab:tabarena_results} reports the corresponding evaluation metrics for all methods.

\paragraph{Improvability and computational cost.}
Improvability measures the remaining performance gap relative to the best result obtained on each task. \textsc{OrionMoE-Small-v1.0} achieves an improvability of $12.77\%$, compared with $12.24\%$ for TabPFN-3, $12.91\%$ for TabICLv2, and Xiaomi-TabLDM. Its improvability is therefore close to that of these competing foundation models, while the strongest models, such as LimiX-2 ($2.06\%$) and Kumo-Tabular-Large ($3.58\%$), leave substantially less room for improvement.

In terms of computational cost, \textsc{OrionMoE-Small-v1.0} requires $4.23$ seconds of support-set processing and $0.94$ seconds of inference per 1,000 samples. Although these costs are substantially lower than those of several computationally intensive foundation models and tuned systems, \textsc{OrionMoE-Small-v1.0} does not offer the lowest latency among comparable methods. In particular, TabPFN-3, TabICLv2, and several other foundation models achieve lower inference times. Consequently, the current advantage of \textsc{OrionMoE-Small-v1.0} lies primarily in its predictive competitiveness relative to several established models, rather than in absolute inference speed.

\begin{table*}[t]
\centering
\small
\setlength{\tabcolsep}{4pt}
\caption{TabArena classifiCation results. Higher is better for Elo, number of wins, and win rate; lower is better for average rank, improvability, and runtime. Training and inference times are normalized per 1,000 samples.}
\label{tab:tabarena_results}
\resizebox{\textwidth}{!}{%
\begin{tabular}{lrrrrrrr}
\toprule
Model & Elo & \#Wins & Avg. Rank $\downarrow$ & Improvability $\downarrow$ & Train / 1K [s] $\downarrow$ & Inference / 1K [s] $\downarrow$ & Win Rate [\%] \\
\midrule
Kumo-Tabular-Large & \textbf{1795} & 7 & \textbf{4.95} & 3.58 & 2.79 & 0.27 & 18.4 \\
LimiX-2 & 1769 & \textbf{10} & 5.47 & \textbf{2.38} & 36.96 & 10.08 & 26.3 \\
Kumo-Tabular-Medium & 1726 & 2 & 6.42 & 5.43 & 1.41 & 0.15 & 5.3 \\
TabPFN-3.5 & 1677 & 3 & 7.68 & 7.21 & 1.85 & 0.52 & 7.9 \\
TabFM & 1583 & 6 & 10.58 & 5.76 & 18.45 & 6.81 & 15.8 \\
Kumo-Tabular-Small & 1558 & 0 & 11.45 & 8.99 & 0.83 & 0.09 & 0.0 \\
TabPFN-3.5-Fast & 1554 & 0 & 11.58 & 8.54 & 0.72 & 0.17 & 0.0 \\
Mitra-v2 & 1533 & 0 & 12.37 & 8.51 & 72.59 & 1.06 & 0.0 \\
Causilo & 1530 & 0 & 12.47 & 9.20 & 0.84 & 0.11 & 0.0 \\
EXAONE-Tabular & 1515 & 0 & 13.08 & 9.36 & 1.53 & 0.49 & 0.0 \\
TabPFN-3 & 1351 & 0 & 20.82 & 12.24 & 1.28 & 0.38 & 0.0 \\
OrionMoE-Small-v1.0 & 1320 & 0 & 22.63 & 12.77 & 4.23 & 0.94 & 0.0 \\
XIAOMI-TabLDM & 1287 & 0 & 24.63 & 12.91 & 1.30 & 0.27 & 0.0 \\
TabICL-v2 & 1277 & 0 & 25.26 & 12.91 & 0.93 & 0.14 & 0.0 \\
RealTabPFN-V2.5 (T+E) & 1244 & 0 & 27.50 & 13.57 & 2046.25 & 8.97 & 0.0 \\
TabDPT-1.3 & 1215 & 0 & 29.58 & 15.27 & 0.62 & 0.17 & 0.0 \\
TabM (T+E) & 1110 & 0 & 37.87 & 17.21 & 2168.86 & 2.35 & 0.0 \\
TabDPT-Turbo & 1090 & 0 & 39.63 & 17.67 & 1.94 & 0.19 & 0.0 \\
TabICL-v1 & 1072 & 0 & 41.21 & 18.15 & 8.68 & 1.74 & 0.0 \\
LightGBM (T+E) & 1071 & 0 & 41.29 & 18.08 & 382.09 & 1.49 & 0.0 \\
LimiX & 1056 & 0 & 42.74 & 17.71 & 18.41 & 4.67 & 0.0 \\
TabDPT (T+E) & 1050 & 0 & 43.29 & 17.72 & 4941.98 & 307.74 & 0.0 \\
CatBoost (T+E) & 1048 & 0 & 43.39 & 17.60 & 1107.00 & 0.33 & 0.0 \\
CatBoost & 1032 & 0 & 44.89 & 18.39 & 5.41 & 0.03 & 0.0 \\
ILTM (T+E) & 1027 & 0 & 45.37 & 18.48 & 12597.42 & 502.45 & 0.0 \\
Mitra-v1 & 1019 & 0 & 46.13 & 20.10 & 121.93 & 2.77 & 0.0 \\
XGBoost (T+E) & 1005 & 0 & 47.45 & 18.86 & 685.87 & 1.45 & 0.0 \\
TabSwift& 996 & 0 & 48.29 & 19.76 & 0.63 & 0.06 & 0.0 \\
TabM & 948 & 0 & 52.89 & 20.33 & 6.85 & 0.25 & 0.0 \\
SAP-RPT-OSS & 948 & 2 & 52.89 & 20.93 & 13.70 & 2.03 & 5.3 \\
TabDPT & 901 & 0 & 57.53 & 21.93 & 49.21 & 43.82 & 0.0 \\
RealMLP & 849 & 0 & 62.39 & 22.06 & 10.48 & 1.71 & 0.0 \\
XGBoost & 835 & 0 & 63.79 & 21.89 & 1.77 & 0.12 & 0.0 \\
RF (T+E) & 791 & 0 & 67.71 & 23.76 & 323.80 & 0.74 & 0.0 \\
LightGBM & 788 & 0 & 68.05 & 22.90 & 1.79 & 0.12 & 0.0 \\
iLTM & 766 & 0 & 69.95 & 25.32 & 262.93 & 64.51 & 0.0 \\
TabSTAR (T+E) & 761 & 0 & 70.36 & 26.94 & 29479.47 & 18.31 & 0.0 \\
OrionMSP & 733 & 0 & 72.71 & 26.99 & 16.32 & 2.86 & 0.0 \\
TabSTAR & 647 & 0 & 79.24 & 32.44 & 439.45 & 5.88 & 0.0 \\
TabFLEX & 626 & 0 & 80.68 & 32.23 & 1.11 & 0.19 & 0.0 \\
RF & 587 & 0 & 83.18 & 30.45 & 0.38 & 0.04 & 0.0 \\
\bottomrule
\end{tabular}
}
\end{table*}

\paragraph{Pairwise performance and ranking consistency.}
The pairwise win-rate matrix in Figure~\ref{fig:tabarena-efficiency} provides a complementary perspective on the Elo ranking by reporting how frequently each method outperforms another across the evaluated datasets.

\textsc{OrionMoE-Small-v1.0} achieves an average rank of $22.63$, compared with $20.82$ for TabPFN-3, $24.63$ for Xiaomi-TabLDM, and $25.26$ for TabICLv2. This ordering is consistent with its overall Elo position and indiCates that its performance is not solely an artifact of the Elo aggregation. However, \textsc{OrionMoE-Small-v1.0} does not attain the best result on any of the 38 evaluated datasets, yielding zero individual dataset wins. This distinction is important: competitive aggregate performance does not necessarily imply state-of-the-art performance on individual tasks. Rather, \textsc{OrionMoE-Small-v1.0} achieves a relatively consistent position among the competing methods without dominating the strongest specialized or larger foundation models.

\subsection{BeyondArena}
\label{app:beyondarena}

\begin{figure*}[htbp]
    \centering
    \begin{subfigure}[t]{0.49\textwidth}
        \centering
        \includegraphics[width=1.05\linewidth]{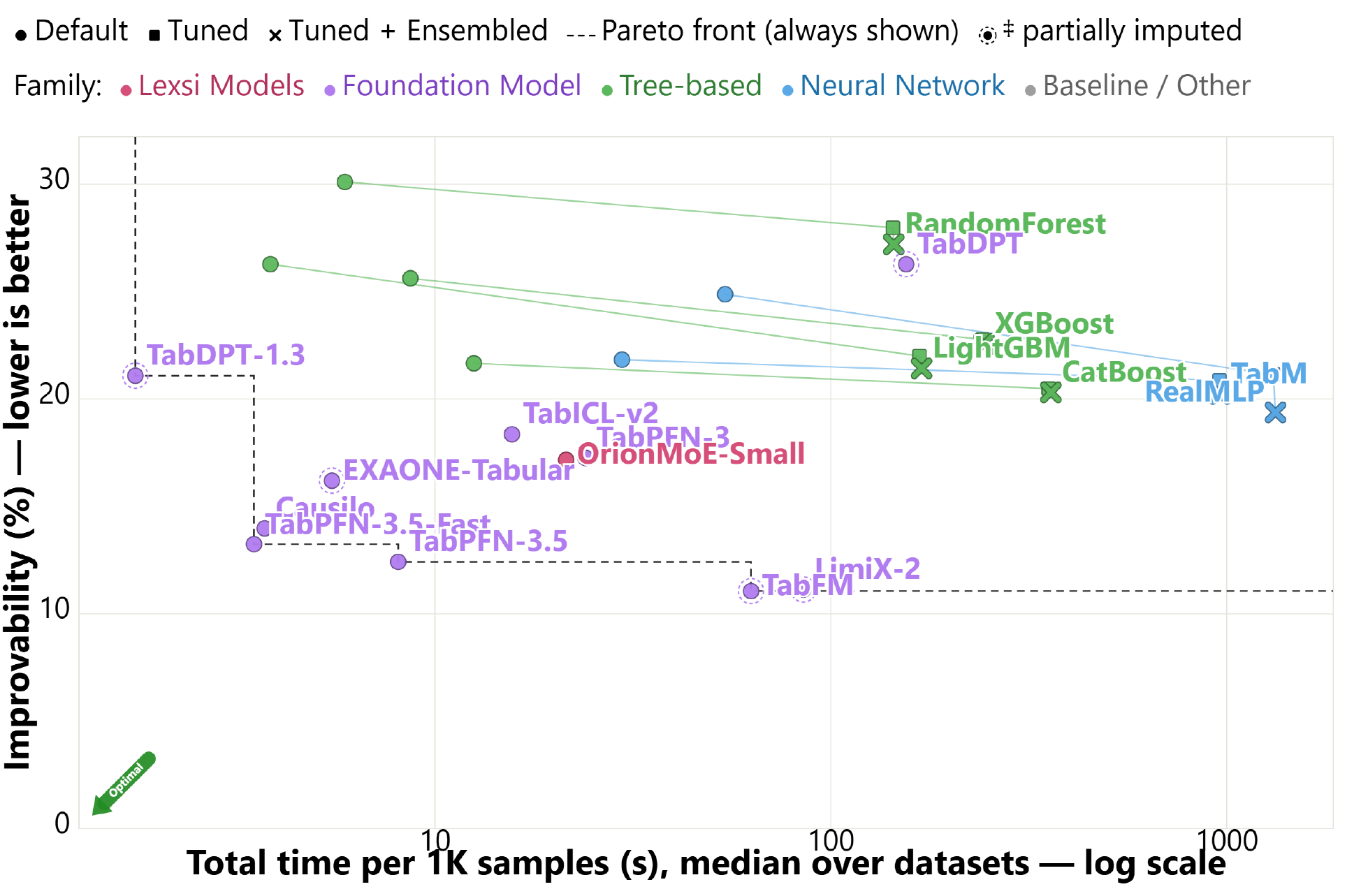}
        \caption{Improvability versus total runtime per 1K samples.}
        \label{fig:beyondarena-improv-total}
    \end{subfigure}
    \hfill
    \begin{subfigure}[t]{0.49\textwidth}
        \centering
        \includegraphics[width=1.1\linewidth]{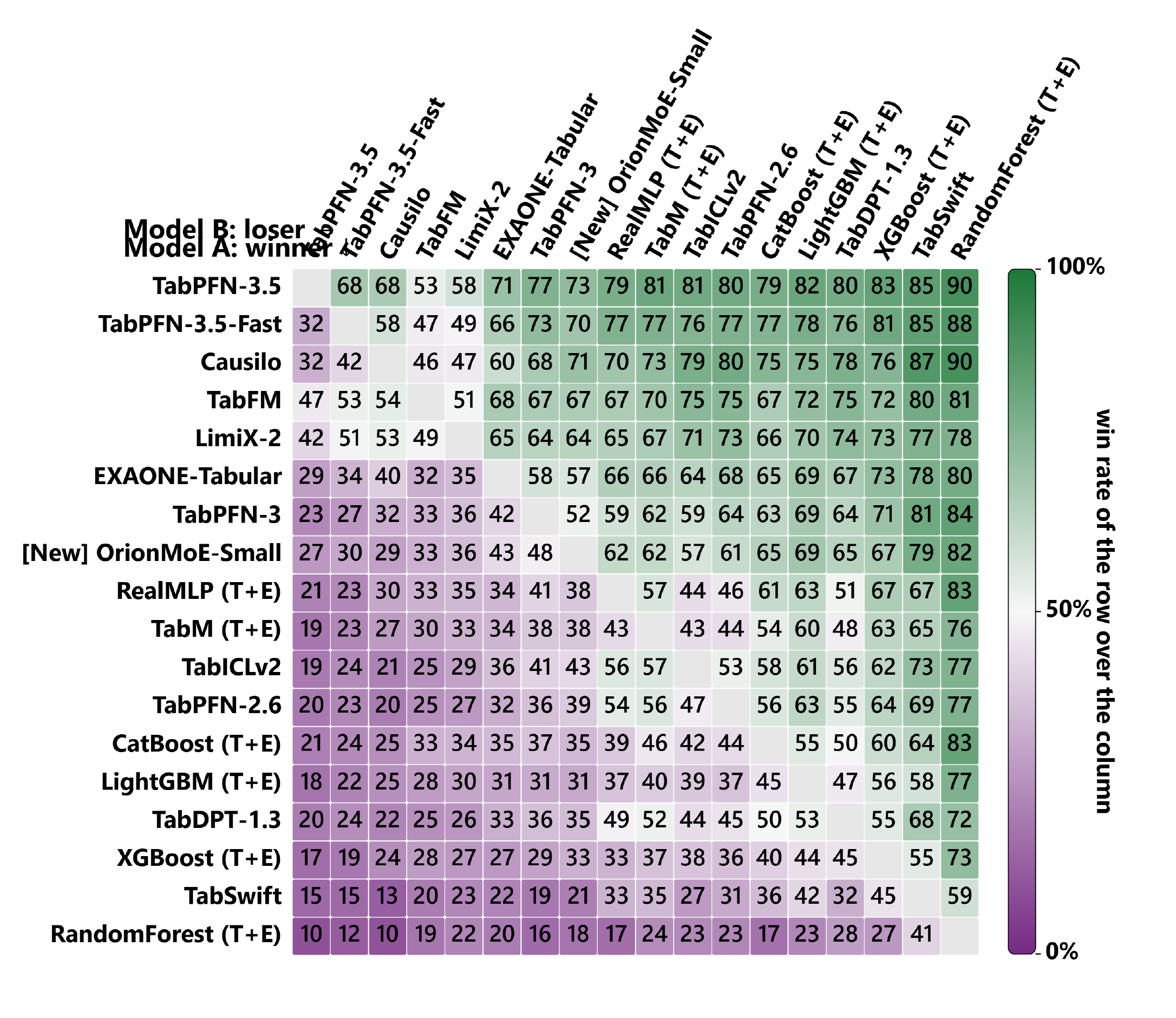}
        \caption{Pairwise win-rate matrix.}
        \label{fig:beyond-winrate-matrix}
    \end{subfigure}
    \caption{Complementary performance analysis on BeyondArena classifiCation tasks. \textbf{(a)} Improvability versus total runtime per 1K samples, capturing the trade-off between additional attainable performance and computational cost. \textbf{(b)} Pairwise win-rate matrix summarizing how consistently each method outperforms competing models across BeyondArena tasks and split regimes.}
    \label{fig:beyondarena-efficiency}
\end{figure*}

We further analyze BeyondArena \cite{beyondarena} using complementary measures of improvability, pairwise performance, and computational cost. These results are reported in Figure~\ref{fig:beyondarena-efficiency} and Table~\ref{tab:beyondarena_results}.

\begin{table*}[t]
\centering
\small
\setlength{\tabcolsep}{4pt}
\caption{BeyondArena classifiCation results over 98 datasets shared by all evaluated methods. Higher is better for Elo, number of wins, and win rate; lower is better for average rank, improvability, and runtime. Training and inference times are normalized per 1,000 samples.}
\label{tab:beyondarena_results}
\resizebox{\textwidth}{!}{%
\begin{tabular}{lrrrrrrr}
\toprule
Model & Elo & \#Wins & Avg. Rank $\downarrow$ & Improvability [\%] $\downarrow$ & Train / 1K [s] $\downarrow$ & Inference / 1K [s] $\downarrow$ & Win Rate [\%] \\
\midrule
TabPFN-3.5 & \textbf{1348} & 19 & \textbf{6.23} & 8.66 & 5.31 & 1.31 & 19.4 \\
TabPFN-3.5-Fast & 1287 & 6 & 7.93 & 9.66 & 2.49 & 0.49 & 6.1 \\
Causilo & 1268 & 4 & 8.52 & 10.27 & 3.20 & 0.24 & 4.1 \\
TabFM & 1222 & \textbf{22.5} & 10.09 & \textbf{8.64} & 41.89 & 9.50 & \textbf{23.0} \\
LimiX-2 & 1188 & 18.5 & 11.35 & 8.75 & 62.33 & 18.46 & 18.9 \\
EXAONE-Tabular & 1176 & 2 & 11.79 & 12.81 & 4.14 & 0.72 & 2.0 \\
TabPFN-3 & 1166 & 1 & 12.21 & 13.82 & 18.11 & 2.98 & 1.0 \\
\textsc{OrionMoE-Small-v1.0} & 1149 & 2 & 12.89 & 13.72 & 13.69 & 2.76 & 2.0 \\
RealMLP (T+E) & 1117 & 1 & 14.24 & 16.14 & 1308.75 & 15.42 & 1.0 \\
TabICLv2 & 1089 & 2 & 15.44 & 15.08 & 11.86 & 2.24 & 2.0 \\
TabM (T+E) & 1076 & 0 & 16.01 & 17.20 & 958.33 & 7.34 & 0.0 \\
CatBoost (T+E) & 1062 & 1 & 16.62 & 17.51 & 354.47 & 1.14 & 1.0 \\
CatBoost (T) & 1049 & 1 & 17.24 & 17.53 & 354.47 & 0.21 & 1.0 \\
RealTabPFN-v2.5 & 1048 & 2 & 17.26 & 17.52 & 4.24 & 0.72 & 2.0 \\
LightGBM (T+E) & 1044 & 2 & 17.46 & 18.30 & 166.76 & 2.19 & 2.0 \\
TabDPT-1.3 & 1041 & 2 & 17.60 & 18.06 & \textbf{1.20} & 0.29 & 2.0 \\
TabM (T) & 1034 & 1 & 17.91 & 17.99 & 958.33 & 1.00 & 1.0 \\
RealMLP (T) & 1028 & 2 & 18.18 & 18.24 & 1308.75 & 1.72 & 2.0 \\
XGBoost (T+E) & 1003 & 2 & 19.33 & 19.62 & 240.99 & 6.55 & 2.0 \\
LightGBM (T) & 1002 & 1 & 19.38 & 18.94 & 166.76 & 0.49 & 1.0 \\
Catboost & 1001 & 1 & 19.41 & 18.80 & 12.42 & 0.17 & 1.0 \\
XGBoost (T) & 979 & 0 & 20.44 & 19.88 & 240.99 & 2.07 & 0.0 \\
TabM & 978 & 1 & 20.46 & 18.78 & 27.38 & 0.91 & 1.0 \\
TabSwift & 935 & 0 & 22.41 & 21.61 & 1.34 & \textbf{0.16} & 0.0 \\
RealMLP & 877 & 1 & 24.99 & 22.34 & 52.60 & 1.57 & 1.0 \\
TabDPT & 876 & 1 & 25.01 & 23.37 & 116.96 & 10.77 & 1.0 \\
XGBoost & 862 & 0 & 25.62 & 22.97 & 6.69 & 1.62 & 0.0 \\
LightGBM & 837 & 0 & 26.62 & 23.82 & 3.60 & 0.19 & 0.0 \\
Random Forest (T+E) & 835 & 0 & 26.70 & 24.32 & 143.10 & 1.60 & 0.0 \\
Random Forest (T) & 795 & 0 & 28.27 & 25.15 & 143.10 & 0.45 & 0.0 \\
Random Forest & 761 & 0 & 29.48 & 27.10 & 5.31 & 0.74 & 0.0 \\
\bottomrule
\end{tabular}
}
\end{table*}

\paragraph{Improvability and runtime.}
\textsc{OrionMoE-Small-v1.0} achieves an improvability of $13.72\%$, closely matching TabPFN-3 ($13.82\%$) and improving upon several competing foundation models, including TabICLv2 ($15.08\%$) and TabDPT-1.3 ($18.06\%$). The strongest methods, notably TabPFN-3.5 and TabFM, obtain substantially lower improvability values of $8.66\%$ and $8.64\%$, respectively.

\textsc{OrionMoE-Small-v1.0} requires $13.69$ seconds of support-set processing and $2.76$ seconds of inference per 1,000 samples. This places its computational cost below several larger or more expensive foundation models, including TabFM, LimiX-2, and several tuned conventional systems. However, faster alternatives remain available, particularly Causilo, EXAONE-Tabular, and TabICLv2. Accordingly, \textsc{OrionMoE-Small-v1.0} occupies an
intermediate region of the performance--runtime trade-off rather than defining its Pareto frontier.

\paragraph{Pairwise comparisons and dataset wins.}
The pairwise win-rate matrix provides additional information about the consistency of the aggregate Elo ranking. \textsc{OrionMoE-Small-v1.0} obtains an average rank of $12.89$, close to TabPFN-3 ($12.21$), and ahead of RealMLP with tuning and ensembling ($14.24$), TabICLv2 ($15.44$), and several conventional learning methods.

Unlike the TabArena evaluation, \textsc{OrionMoE-Small-v1.0} achieves the highest predictive performance on two BeyondArena datasets, corresponding to a dataset win rate of approximately $2.0\%$. Although this remains below
the number of wins attained by the leading models, it indiCates that \textsc{OrionMoE-Small-v1.0} can outperform the complete comparison set on individual tasks outside the conventional IID setting.

The pairwise analysis also highlights the diversity of the benchmark. No single method dominates all evaluation regimes, and the relative strength of tuned conventional models increases on grouped, temporal, and
large-scale datasets. In this context, \textsc{OrionMoE-Small-v1.0} remains competitive with several foundation-model baselines but does not consistently displace the strongest task-specific systems.

Overall, these complementary metrics support the conclusions of the main BeyondArena analysis. \textsc{OrionMoE-Small-v1.0} achieves competitive aggregate performance and a moderate improvability gap, while the remaining differences to the leading methods reflect both predictive limitations and computational trade-offs that vary substantially across dataset regimes.


\subsection{Common-Task Benchmark Results}
\label{app:common_benchmarks}

We complement the Elo-based comparisons in the main report with mean classifiCation accuracy and average model rank. All metrics are computed over the common-task subsets used in the corresponding Elo evaluations, ensuring that participating models are evaluated on identical datasets.

\begin{figure}[htbp]
    \centering
    \includegraphics[width=\linewidth]{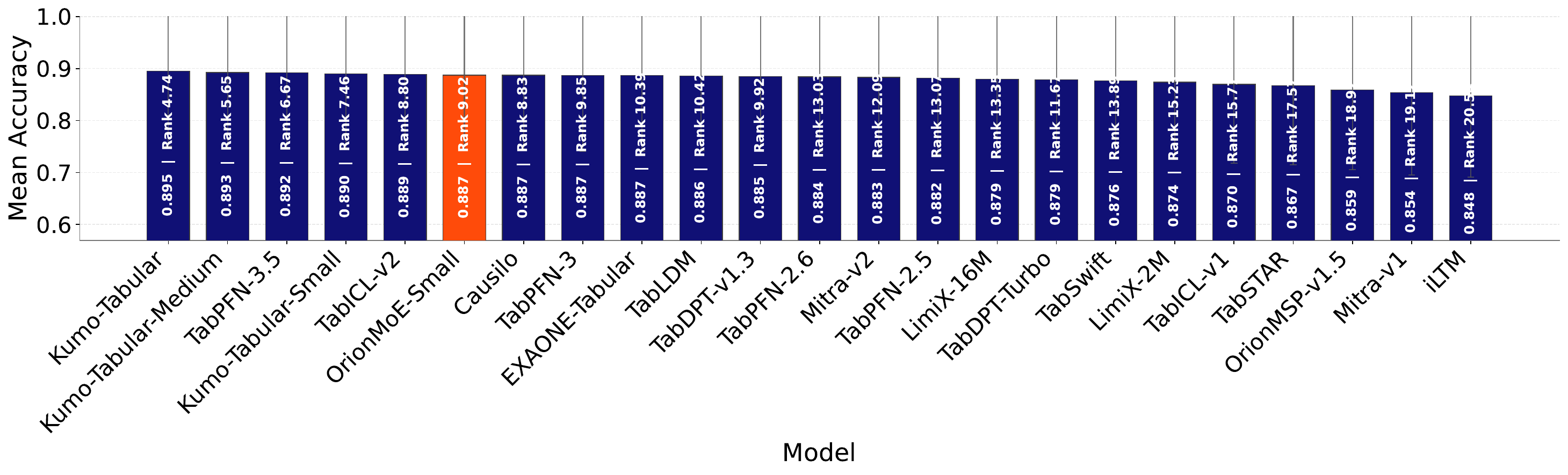}
    \caption{Mean accuracy on the common-task subset of OpenML-CC18. Each bar reports the average accuracy of a model across the shared set of tasks, with the corresponding overall rank shown inside the bar.}
    \label{fig:openml_accuracy}
\end{figure}
    
\paragraph{OpenML-CC18.}
Figure~\ref{fig:openml_accuracy} reports mean accuracy and average rank across 48 common OpenML-CC18 tasks. \textsc{OrionMoE-Small-v1.0} achieves a mean accuracy of $0.887$ and an average rank of $9.02$, placing it among the leading models in this evaluation. Its accuracy is close to TabPFN-3.5 ($0.892$) and TabICLv2 ($0.889$), and matches the displayed mean accuracy of Causilo, TabPFN-3 and EXAONE-Tabular. The average-rank comparison provides additional context: \textsc{OrionMoE-Small-v1.0} remains competitive across datasets, although TabPFN-3.5, Causilo, and TabICLv2 obtain somewhat stronger average ranks. Together, these results complement the Elo ranking and show that the differences among several leading models are small in absolute mean accuracy.

\begin{figure*}[htbp]
    \centering
    \begin{subfigure}[t]{0.49\textwidth}
        \centering
        \includegraphics[width=1.\linewidth]{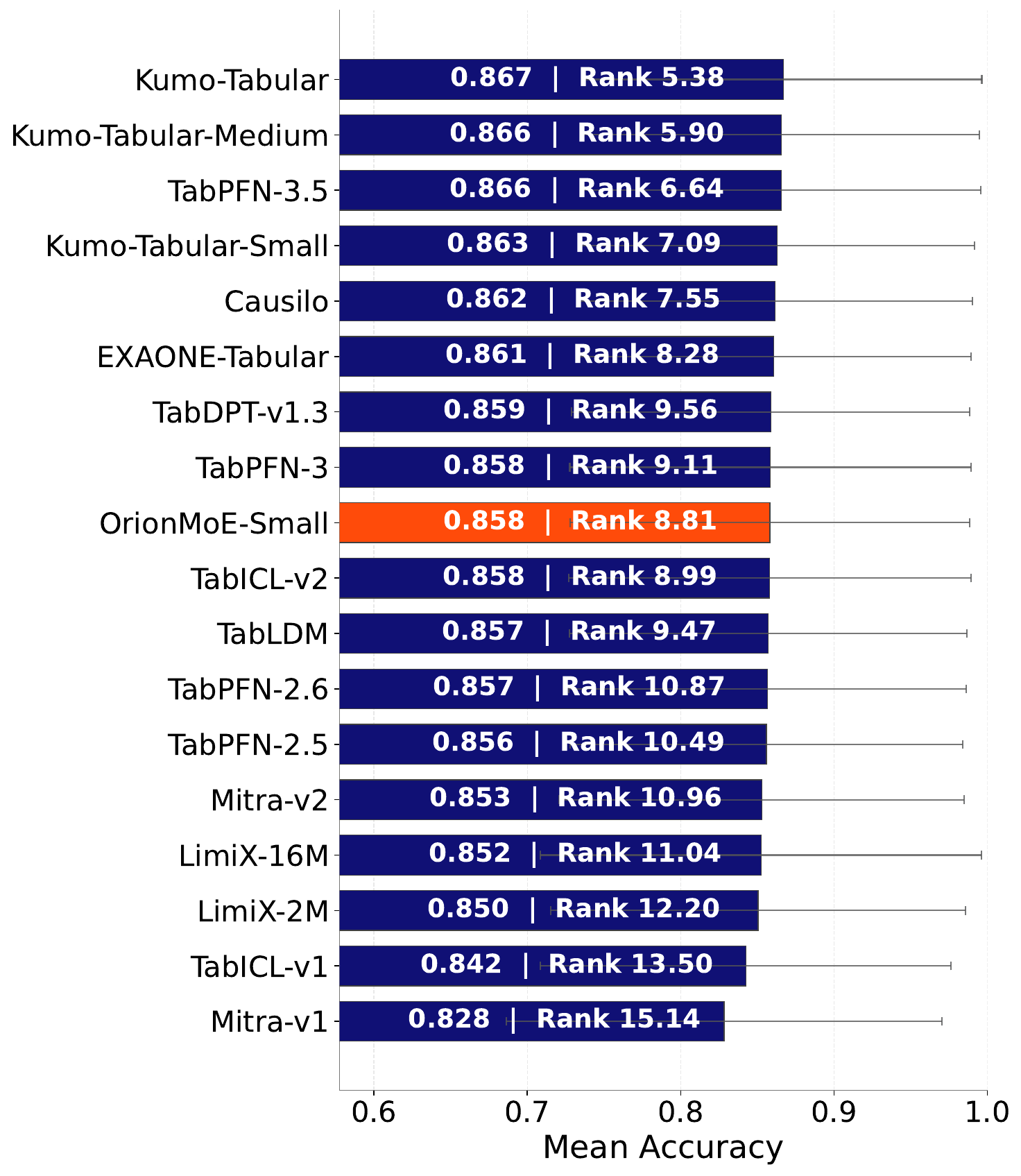}
        \caption{Mean accuracy on common TALENT tasks.}
        \label{fig:talent_accuracy}
    \end{subfigure}
    \hfill
    \begin{subfigure}[t]{0.49\textwidth}
        \centering
        \includegraphics[width=1\linewidth]{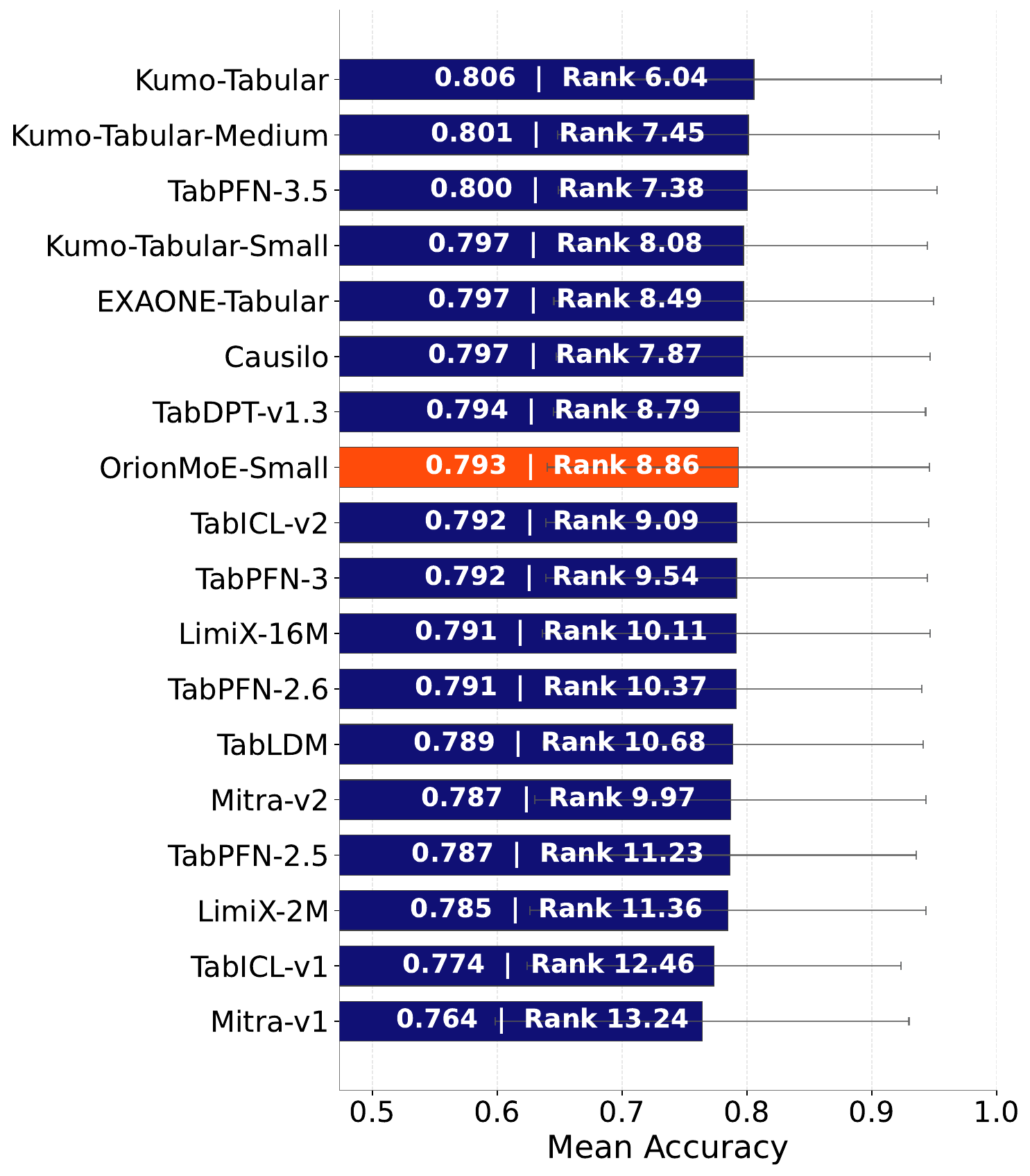}
        \caption{Mean accuracy on common BCCO tasks.}
        \label{fig:bcco_accuracy}
    \end{subfigure}
    \caption{Mean accuracy on the common-task subsets of TALENT and BCCO. \textbf{(a)} TALENT and \textbf{(b)} BCCO. Bars report the average accuracy over the shared task set, with the corresponding model rank shown inside each bar.}
    \label{fig:benchmarks_accuracy}
\end{figure*}

\paragraph{TALENT.}
Across the $148$ common TALENT datasets, \textsc{OrionMoE-Small-v1.0} obtains a mean accuracy of $0.852$ and an average rank of $8.81$, as shown in Figure~\ref{fig:talent_accuracy}. Its mean accuracy matches TabPFN-3 and TabICLv2 at the displayed precision, while its average rank is slightly better than both methods ($9.11$ and $8.99$, respectively). \textsc{OrionMoE-Small-v1.0} also exceeds TabLDM ($0.851$) and several earlier foundation models in mean accuracy. Although TabPFN-3.5 remains the strongest method with $0.859$ accuracy, the difference is less than one percentage point. These results indiCate that \textsc{OrionMoE-Small-v1.0} maintains competitive accuracy across a heterogeneous collection of tabular classifiCation problems.

\paragraph{BCCO.}
Figure~\ref{fig:bcco_accuracy} shows that \textsc{OrionMoE-Small-v1.0} achieves a mean accuracy of $0.793$ and an average rank of $8.86$ over the 86 common BCCO datasets. Its mean accuracy is close to TabDPT-v1.3 ($0.794$), Causilo ($0.797$), and EXAONE-Tabular ($0.797$), while remaining within $0.013$ absolute accuracy of the leading Kumo-Tabular-Large model ($0.806$). \textsc{OrionMoE-Small-v1.0} also outperforms TabICLv2, TabPFN-3, TabLDM, and several other foundation models in mean accuracy. This provides additional evidence that its competitive position persists across the different benchmark collections.

Overall, the mean-accuracy results are consistent with the Elo-based evaluation: \textsc{OrionMoE-Small-v1.0} remains within a relatively narrow accuracy range of the leading methods, while its average rank confirms competitive performance across the shared evaluation tasks.

\subsection{Full-Benchmark Accuracy and Ranking}
\label{app:full_benchmark_accuracy}

We additionally report mean accuracy and average rank for models completing the entire OpenML-CC18, TALENT, and BCCO benchmark suites. These results complement the full-benchmark Elo comparisons presented in Section~\ref{sec:full_benchmarks}.

\begin{figure*}[htbp]
    \centering
    \begin{subfigure}[t]{0.49\textwidth}
        \centering
        \includegraphics[width=1.\linewidth]{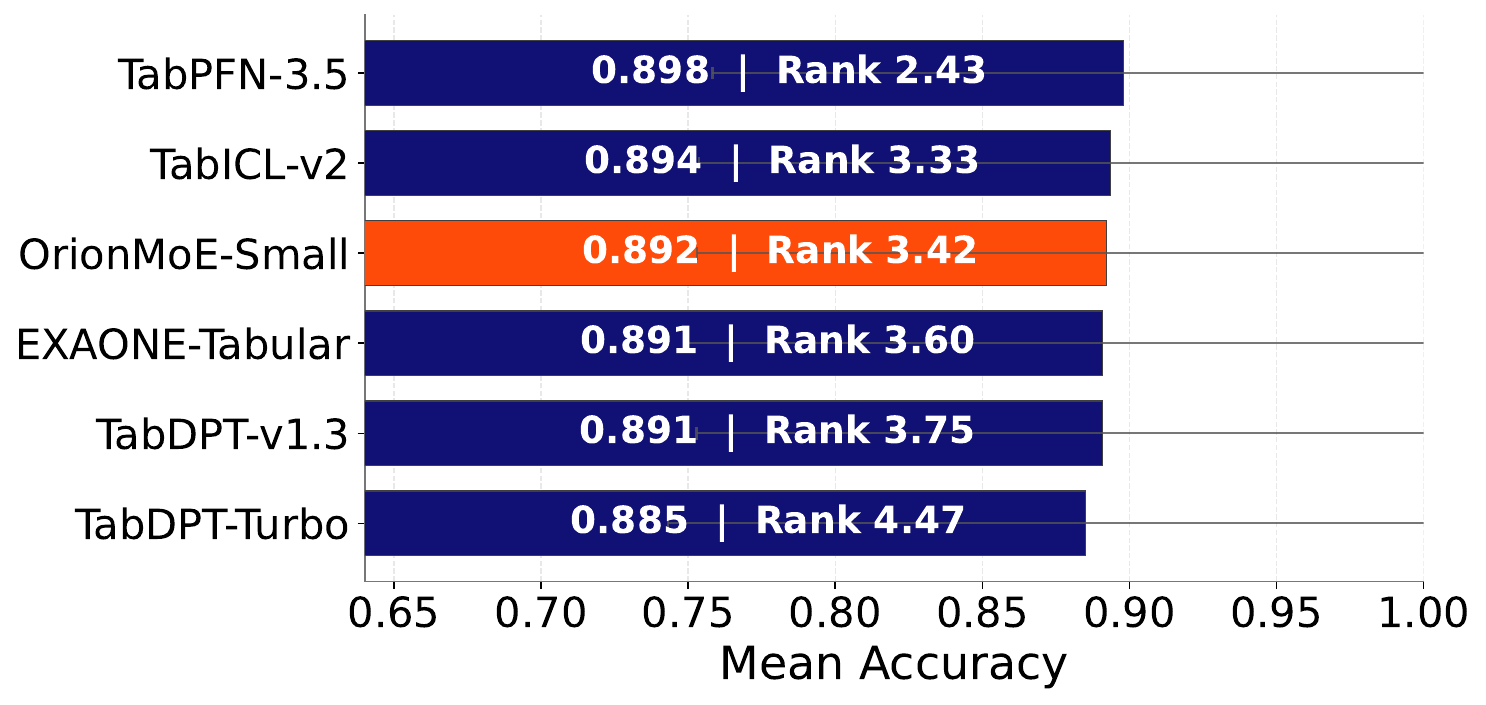}
        \caption{OpenML-CC18: models completing the full benchmark}
        \label{fig:talent_full_accuracy}
    \end{subfigure}
    \hfill
    \begin{subfigure}[t]{0.49\textwidth}
        \centering
        \includegraphics[width=1\linewidth]{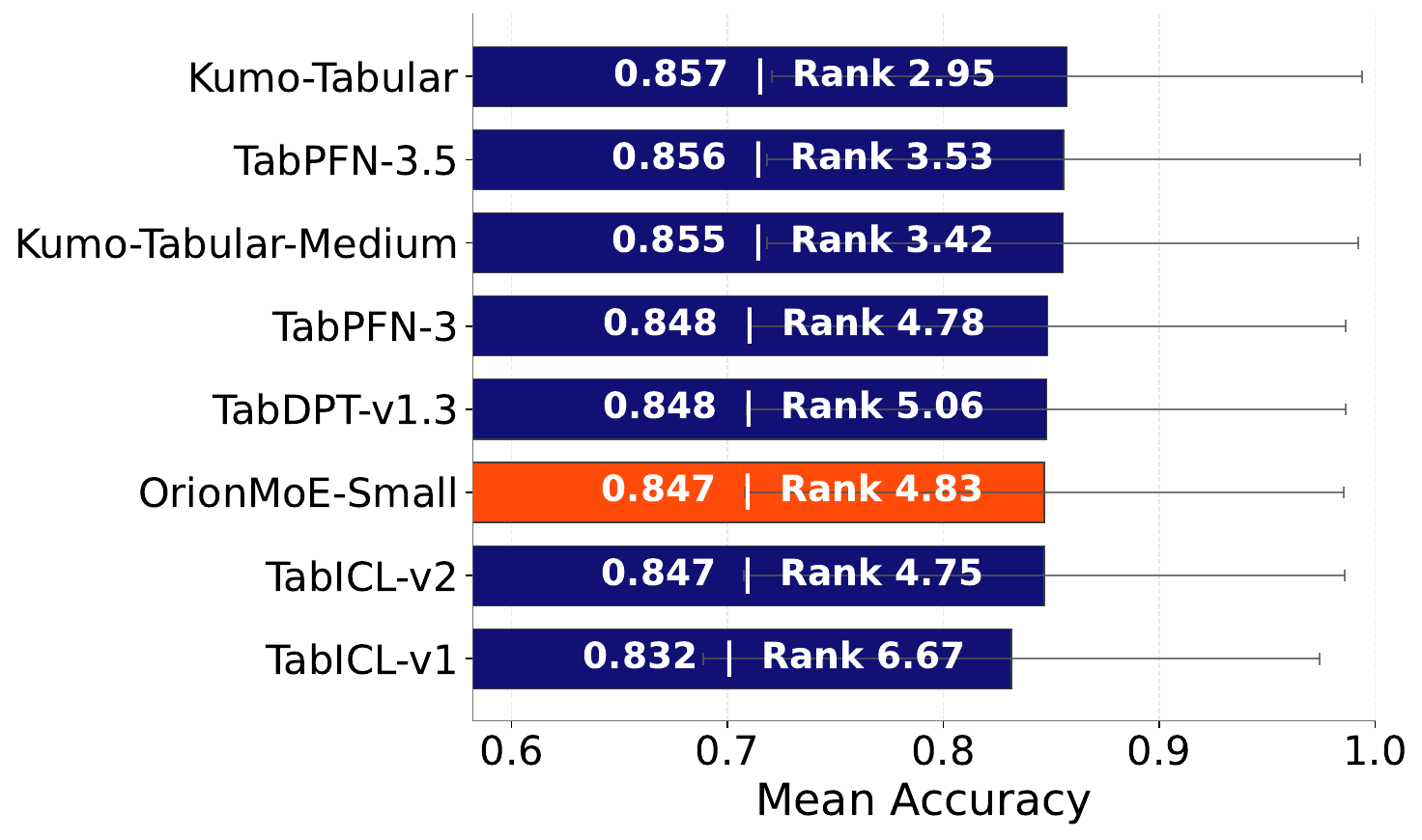}
        \caption{TALENT: models completing the full benchmark.}
        \label{fig:bcco_full_accuracy}
    \end{subfigure}
    \caption{Mean accuracy for models that completed the full benchmark suite. \textbf{(a)} OpenML-CC18 and \textbf{(b)} TALENT. Bars report the average accuracy across all benchmark tasks completed by every included model, with the corresponding rank shown inside each bar.}
    \label{fig:benchmarks_full_accuracy}
\end{figure*}

Figure~\ref{fig:benchmarks_full_accuracy} summarizes performance on OpenML-CC18 and TALENT, while Figure~\ref{fig:bcco_full_accuracy} reports the corresponding BCCO results.

\begin{figure}[htbp]
    \centering
    \includegraphics[width=\linewidth]{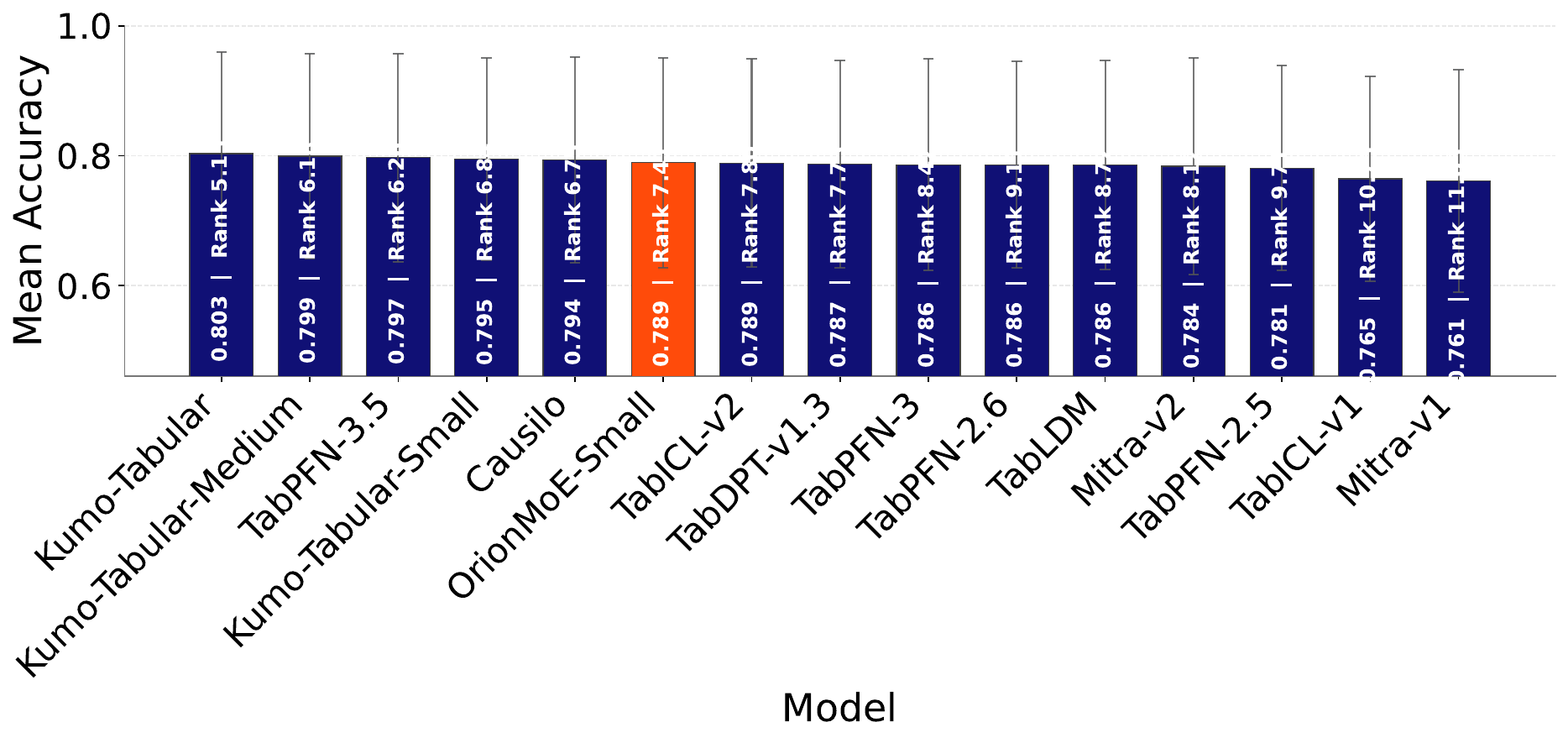}
    \caption{Mean accuracy on the full BCCO benchmark for models that completed all tasks. Each bar reports the average accuracy over the complete benchmark, with the corresponding model rank shown inside the bar.}
    \label{fig:bcc_full_accuracy}
\end{figure}

On the complete OpenML-CC18 benchmark, \textsc{OrionMoE-Small-v1.0} achieves a mean accuracy of $0.892$ with an average rank of $3.42$. Its accuracy is close to TabPFN-3.5 ($0.898$) and TabICLv2 ($0.894$), while slightly exceeding EXAONE-Tabular and TabDPT-v1.3 at the reported precision. Importantly, the average-rank results place \textsc{OrionMoE-Small-v1.0} close to the strongest participating models, indicating competitive performance across the complete 72-task collection.

On TALENT, \textsc{OrionMoE-Small-v1.0} obtains a mean accuracy of $0.847$ and an average rank of $4.83$ across all 181 datasets. Its mean accuracy matches TabICLv2 at the reported precision and is within $0.001$ of TabPFN-3 and TabDPT-v1.3. The leading model, Kumo-Tabular-Large, reaches $0.857$, leaving an absolute difference of $0.010$.

Finally, \textsc{OrionMoE-Small-v1.0} achieves $0.789$ mean accuracy and an average rank of $7.4$ on the full BCCO benchmark. While TabPFN-3.5 ($0.797$) and Causilo ($0.794$) obtain higher accuracy, \textsc{OrionMoE-Small-v1.0} matches TabICLv2 at the displayed precision and exceeds TabDPT-v1.3, TabPFN-3, TabLDM, and several other models. Its average rank is also slightly better than that of TabICLv2 ($7.84$).

These complementary metrics confirm that \textsc{OrionMoE-Small-v1.0} remains competitive across the full benchmark suites. In particular, the relatively small differences in mean accuracy between several of the leading methods indiCate that their Elo orderings should be interpreted alongside average ranks and dataset-level performance rather than as absolute measures of predictive superiority.

\subsection{Performance Across Class Imbalance Regimes}
\label{app:class_imbalance}

To further characterize \textsc{OrionMoE-Small-v1.0} across dataset properties, we analyze its performance as a function of class imbalance using datasets from OpenML-CC18, TALENT, and BCCO. We consider three Categories: near-balanced datasets, moderately imbalanced datasets, and severely imbalanced datasets, with the latter defined by a class imbalance ratio of at least $10$.

\begin{figure}[htbp]
    \centering
    \includegraphics[width=\linewidth]{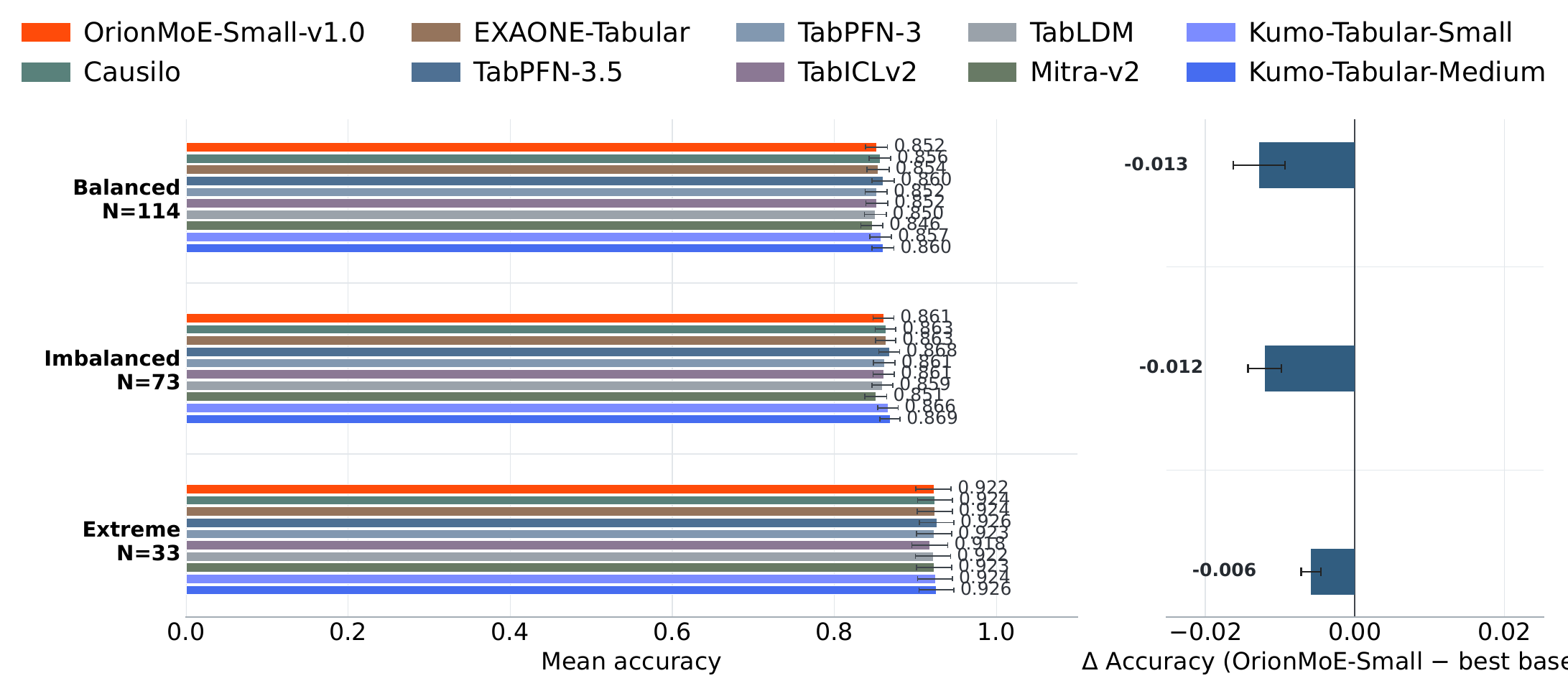}
    \caption{Performance across datasets grouped by class imbalance ratio, after merging OpenML-CC18, TALENT, and BCCO. The top panel reports mean accuracy for each model within three imbalance regimes: near-balanced, moderate imbalance, and severe imbalance ($\geq 10\times$). The bottom panel reports the accuracy difference between \textsc{OrionMoE-Small-v1.0} and the strongest competing baseline within each regime, where values closer to zero indiCate stronger relative performance. Overall, \textsc{OrionMoE-Small-v1.0} remains competitive across all imbalance levels, with the smallest gap observed in the most severely imbalanced regime.}
    \label{fig:class-imbalance}
\end{figure}

Figure~\ref{fig:class-imbalance} compares mean accuracy across eight recent tabular foundation models. The lower panel reports the mean per-dataset accuracy difference between \textsc{OrionMoE-Small-v1.0} and the strongest competing baseline.

An interesting trend emerges in the relative performance of OrionMoE-Small. Its mean accuracy gap to the strongest competing model decreases from $-0.013$ on near-balanced datasets to $-0.012$ under moderate imbalance and only $-0.006$ under severe imbalance. Thus, \textsc{OrionMoE-Small-v1.0} becomes progressively closer to the strongest competing methods as the class distribution becomes more skewed.

This pattern complements the TabArena regime-specific evaluation, where \textsc{OrionMoE-Small-v1.0} also demonstrates competitive performance on imbalanced classifiCation tasks. However, the two analyses are not directly equivalent: TabArena uses independently computed Elo ratings, whereas the present comparison measures paired accuracy differences across datasets.

The decreasing relative gap suggests that class imbalance does not systematically disadvantage \textsc{OrionMoE-Small-v1.0} relative to the evaluated foundation models. Nevertheless, accuracy alone does not establish improved recognition of minority classes, and the smaller number of severely imbalanced datasets limits the strength of this conclusion. Additional class-sensitive metrics would be necessary to determine whether the observed competitiveness extends to minority-class predictive performance.

\section{Datasets}
\label{app:datasets}

\subsection{OpenML-CC18}
\label{app:data-openml}
\begin{longtable}{lrrrrr}
\caption{OpenML-CC18 datasets used in the common 48-task evaluation.  For each dataset, we report the OpenML task ID, context size,  query size, number of input features, and number of classes. Context and query sizes are reported as averages across evaluation folds and can therefore be non-integer}
\label{tab:openml_cc18_tasks} \\
\toprule
Dataset & Task ID & Context Size & Query Size & \# Features & \# Classes \\
\midrule
\endfirsthead

\multicolumn{6}{c}%
{{\tablename\ \thetable{} -- continued from previous page}} \\
\toprule
Dataset & Task ID & Context Size & Query Size & \# Features & \# Classes \\
\midrule
\endhead

\midrule
\multicolumn{6}{r}{{Continued on next page}} \\
\endfoot

\bottomrule
\endlastfoot

kr-vs-kp                              & 3      & 2876.4  & 319.6  & 36  & 2  \\
balance-scale                         & 11     & 562.5   & 62.5   & 4   & 3  \\
mfeat-factors                         & 12     & 1800.0  & 200.0  & 216 & 10 \\
mfeat-fourier                         & 14     & 1800.0  & 200.0  & 76  & 10 \\
mfeat-karhunen                        & 16     & 1800.0  & 200.0  & 64  & 10 \\
mfeat-morphological                   & 18     & 1800.0  & 200.0  & 6   & 10 \\
mfeat-zernike                         & 22     & 1800.0  & 200.0  & 47  & 10 \\
cmc                                   & 23     & 1325.7  & 147.3  & 9   & 3  \\
optdigits                             & 28     & 5058.0  & 562.0  & 64  & 10 \\
credit-g                              & 31     & 900.0   & 100.0  & 20  & 2  \\
pendigits                             & 32     & 9892.8  & 1099.2 & 16  & 10 \\
diabetes                              & 37     & 691.2   & 76.8   & 8   & 2  \\
spambase                              & 43     & 4140.9  & 460.1  & 57  & 2  \\
splice                                & 45     & 2871.0  & 319.0  & 60  & 3  \\
tic-tac-toe                           & 49     & 862.2   & 95.8   & 9   & 2  \\
vehicle                               & 53     & 761.4   & 84.6   & 18  & 4  \\
electricity                           & 219    & 40780.8 & 4531.2 & 8   & 2  \\
satimage                              & 2074   & 5787.0  & 643.0  & 36  & 6  \\
analcatdata\_authorship               & 3549   & 756.9   & 84.1   & 70  & 4  \\
analcatdata\_dmft                     & 3560   & 717.3   & 79.7   & 4   & 6  \\
pc4                                   & 3902   & 1312.2  & 145.8  & 37  & 2  \\
pc3                                   & 3903   & 1406.7  & 156.3  & 37  & 2  \\
kc2                                   & 3913   & 469.8   & 52.2   & 21  & 2  \\
kc1                                   & 3917   & 1898.1  & 210.9  & 21  & 2  \\
pc1                                   & 3918   & 998.1   & 110.9  & 21  & 2  \\
adult                                 & 7592   & 43957.8 & 4884.2 & 14  & 2  \\
wdbc                                  & 9946   & 512.1   & 56.9   & 30  & 2  \\
phoneme                               & 9952   & 4863.6  & 540.4  & 5   & 2  \\
qsar-biodeg                           & 9957   & 949.5   & 105.5  & 41  & 2  \\
wall-robot-navigation                 & 9960   & 4910.4  & 545.6  & 24  & 4  \\
semeion                               & 9964   & 1433.7  & 159.3  & 256 & 10 \\
ilpd                                  & 9971   & 524.7   & 58.3   & 10  & 2  \\
madelon                               & 9976   & 2340.0  & 260.0  & 500 & 2  \\
ozone-level-8hr                       & 9978   & 2280.6  & 253.4  & 72  & 2  \\
first-order-theorem-proving           & 9985   & 5506.2  & 611.8  & 51  & 6  \\
banknote-authentication               & 10093  & 1234.8  & 137.2  & 4   & 2  \\
blood-transfusion-service-center      & 10101  & 673.2   & 74.8   & 4   & 2  \\
PhishingWebsites                      & 14952  & 9949.5  & 1105.5 & 30  & 2  \\
bank-marketing                        & 14965  & 40689.9 & 4521.1 & 16  & 2  \\
GesturePhaseSegmentationProcessed     & 14969  & 8885.7  & 987.3  & 32  & 5  \\
dresses-sales                         & 125920 & 450.0   & 50.0   & 12  & 2  \\
steel-plates-fault                    & 146817 & 1746.9  & 194.1  & 27  & 7  \\
climate-model-simulation-crashes      & 146819 & 486.0   & 54.0   & 18  & 2  \\
wilt                                  & 146820 & 4355.1  & 483.9  & 5   & 2  \\
car                                   & 146821 & 1555.2  & 172.8  & 6   & 4  \\
segment                               & 146822 & 2079.0  & 231.0  & 16  & 7  \\
mfeat-pixel                           & 146824 & 1800.0  & 200.0  & 240 & 10 \\
jungle\_chess\_2pcs\_raw\_endgame\_complete
                                      & 167119 & 40337.1 & 4481.9 & 6   & 3  \\

\end{longtable}

\subsection{TALENT}
\label{app:data-talent}

\begin{longtable}{p{0.52\textwidth}rrrr}
\caption{TALENT datasets used in the common 148-dataset evaluation. Context size denotes the number of training examples available to the model at inference time, while query size corresponds to the held-out test set size.}
\label{tab:talent_datasets} \\
\toprule
Dataset & Context Size & Query Size & \# Features & \# Classes \\
\midrule
\endfirsthead

\multicolumn{5}{c}%
{{\tablename\ \thetable{} -- continued from previous page}} \\
\toprule
Dataset & Context Size & Query Size & \# Features & \# Classes \\
\midrule
\endhead

\midrule
\multicolumn{5}{r}{{Continued on next page}} \\
\endfoot

\bottomrule
\endlastfoot

Amazon\_employee\_access & 26215 & 6554 & 7 & 2 \\
BLE\_RSSI\_dataset\_for\_Indoor\_localization & 7987 & 1997 & 3 & 3 \\
BNG(breast-w) & 31492 & 7874 & 9 & 2 \\
BNG(cmc) & 44236 & 11060 & 9 & 3 \\
BNG(tic-tac-toe) & 31492 & 7874 & 9 & 2 \\
Bank\_Customer\_Churn\_Dataset & 8000 & 2000 & 10 & 2 \\
Basketball\_c & 1072 & 268 & 11 & 2 \\
California-Housing-Classification & 16512 & 4128 & 8 & 2 \\
Click\_prediction\_small & 31958 & 7990 & 3 & 2 \\
Customer\_Personality\_Analysis & 1792 & 448 & 24 & 2 \\
Diabetic\_Retinopathy\_Debrecen & 920 & 231 & 19 & 2 \\
E-CommereShippingData & 8799 & 2200 & 10 & 2 \\
Employee & 3722 & 931 & 8 & 2 \\
FICO-HELOC-cleaned & 7896 & 1975 & 23 & 2 \\
FOREX\_audcad-day-High & 1467 & 367 & 10 & 2 \\
FOREX\_audchf-day-High & 1466 & 367 & 10 & 2 \\
FOREX\_audjpy-day-High & 1465 & 367 & 10 & 2 \\
FOREX\_audjpy-hour-High & 35060 & 8765 & 10 & 2 \\
FOREX\_audsgd-hour-High & 35060 & 8765 & 10 & 2 \\
FOREX\_audusd-hour-High & 35060 & 8765 & 10 & 2 \\
FOREX\_cadjpy-day-High & 1467 & 367 & 10 & 2 \\
FOREX\_cadjpy-hour-High & 35060 & 8765 & 10 & 2 \\
Firm-Teacher\_Clave-Direction\_Classification & 8640 & 2160 & 16 & 4 \\
Fitness\_Club\_c & 1200 & 300 & 6 & 2 \\
GAMETES\_Epistasis\_2-Way\_20atts\_0.1H\_EDM-1\_1 & 1280 & 320 & 20 & 2 \\
GAMETES\_Heterogeneity\_20atts\_1600\_Het\_0.4\_0.2\_50\_EDM-2\_001 & 1280 & 320 & 20 & 2 \\
Gender\_Gap\_in\_Spanish\_WP & 3796 & 950 & 13 & 3 \\
GesturePhaseSegmentationProcessed & 7898 & 1975 & 32 & 5 \\
HR\_Analytics\_Job\_Change\_of\_Data\_Scientists & 15326 & 3832 & 13 & 2 \\
IBM\_HR\_Analytics\_Employee\_Attrition\_and\_Performance & 1176 & 294 & 31 & 2 \\
INNHotelsGroup & 29020 & 7255 & 17 & 2 \\
Indian\_pines & 7315 & 1829 & 220 & 8 \\
JapaneseVowels & 7968 & 1993 & 14 & 9 \\
KDDCup09\_upselling & 4102 & 1026 & 49 & 2 \\
MIC & 1319 & 330 & 104 & 2 \\
MagicTelescope & 15216 & 3804 & 9 & 2 \\
Marketing\_Campaign & 1792 & 448 & 27 & 2 \\
Mobile\_Price\_Classification & 1600 & 400 & 20 & 4 \\
National\_Health\_and\_Nutrition\_Health\_Survey & 1822 & 456 & 7 & 2 \\
PhishingWebsites & 8844 & 2211 & 30 & 2 \\
PieChart3 & 861 & 216 & 37 & 2 \\
Pima\_Indians\_Diabetes\_Database & 614 & 154 & 8 & 2 \\
PizzaCutter3 & 834 & 209 & 37 & 2 \\
Pumpkin\_Seeds & 2000 & 500 & 12 & 2 \\
QSAR\_biodegradation & 843 & 211 & 41 & 2 \\
Satellite & 4080 & 1020 & 36 & 2 \\
Telecom\_Churn\_Dataset & 2666 & 667 & 17 & 2 \\
Water\_Quality\_and\_Potability & 2620 & 656 & 8 & 2 \\
Wilt & 3856 & 965 & 5 & 2 \\
abalone & 3341 & 836 & 8 & 3 \\
ada & 3317 & 830 & 48 & 2 \\
ada\_agnostic & 3649 & 913 & 48 & 2 \\
ada\_prior & 3649 & 913 & 14 & 2 \\
airlines\_seed\_0\_nrows\_2000\_nclasses\_10\_ncols\_100\_stratify\_True & 1600 & 400 & 7 & 2 \\
allbp & 3017 & 755 & 29 & 3 \\
allrep & 3017 & 755 & 29 & 4 \\
analcatdata\_authorship & 672 & 169 & 69 & 4 \\
artificial-characters & 8174 & 2044 & 7 & 10 \\
autoUniv-au4-2500 & 2000 & 500 & 100 & 3 \\
autoUniv-au7-1100 & 880 & 220 & 12 & 5 \\
bank & 36168 & 9043 & 16 & 2 \\
banknote\_authentication & 1097 & 275 & 4 & 2 \\
baseball & 1072 & 268 & 16 & 3 \\
car-evaluation & 1382 & 346 & 21 & 4 \\
churn & 4000 & 1000 & 20 & 2 \\
cmc & 1178 & 295 & 9 & 3 \\
company\_bankruptcy\_prediction & 5455 & 1364 & 95 & 2 \\
compass & 13315 & 3329 & 17 & 2 \\
contraceptive\_method\_choice & 1178 & 295 & 9 & 3 \\
credit & 13371 & 3343 & 10 & 2 \\
delta\_ailerons & 5703 & 1426 & 5 & 2 \\
dis & 3017 & 755 & 29 & 2 \\
dna & 2548 & 638 & 180 & 3 \\
drug\_consumption & 1507 & 377 & 12 & 7 \\
dry\_bean\_dataset & 10888 & 2723 & 16 & 7 \\
eeg-eye-state & 11984 & 2996 & 14 & 2 \\
electricity & 36249 & 9063 & 8 & 2 \\
estimation\_of\_obesity\_levels & 1688 & 423 & 16 & 7 \\
eye\_movements & 8748 & 2188 & 27 & 3 \\
eye\_movements\_bin & 6086 & 1522 & 20 & 2 \\
first-order-theorem-proving & 4894 & 1224 & 51 & 6 \\
gas-drift & 11128 & 2782 & 128 & 6 \\
gina\_agnostic & 2774 & 694 & 970 & 2 \\
golf\_play\_dataset\_extended & 876 & 219 & 9 & 2 \\
heloc & 8000 & 2000 & 22 & 2 \\
hill-valley & 969 & 243 & 100 & 2 \\
house\_16H & 10790 & 2698 & 16 & 2 \\
htru & 14318 & 3580 & 8 & 2 \\
ibm-employee-performance & 1176 & 294 & 30 & 2 \\
in\_vehicle\_coupon\_recommendation & 10147 & 2537 & 21 & 2 \\
jm1 & 8708 & 2177 & 21 & 2 \\
jungle\_chess\_2pcs\_raw\_endgame\_complete & 35855 & 8964 & 6 & 3 \\
kc1 & 1687 & 422 & 21 & 2 \\
kdd\_ipums\_la\_97-small & 4150 & 1038 & 20 & 2 \\
led24 & 2560 & 640 & 24 & 10 \\
led7 & 2560 & 640 & 7 & 10 \\
mammography & 8946 & 2237 & 6 & 2 \\
maternal\_health\_risk & 811 & 203 & 6 & 3 \\
mfeat-factors & 1600 & 400 & 216 & 10 \\
mfeat-fourier & 1600 & 400 & 76 & 10 \\
mfeat-karhunen & 1600 & 400 & 64 & 10 \\
mfeat-morphological & 1600 & 400 & 6 & 10 \\
mfeat-pixel & 1600 & 400 & 240 & 10 \\
mfeat-zernike & 1600 & 400 & 47 & 10 \\
mice\_protein\_expression & 864 & 216 & 75 & 8 \\
microaggregation2 & 16000 & 4000 & 20 & 5 \\
mobile\_c36\_oversampling & 41408 & 10352 & 6 & 2 \\
mozilla4 & 12436 & 3109 & 4 & 2 \\
national-longitudinal-survey-binary & 3926 & 982 & 16 & 2 \\
online\_shoppers & 9864 & 2466 & 14 & 2 \\
optdigits & 4496 & 1124 & 64 & 10 \\
ozone-level-8hr & 2027 & 507 & 72 & 2 \\
page-blocks & 4378 & 1095 & 10 & 5 \\
pc1 & 887 & 222 & 21 & 2 \\
pc3 & 1250 & 313 & 37 & 2 \\
pc4 & 1166 & 292 & 37 & 2 \\
pendigits & 8793 & 2199 & 16 & 10 \\
phoneme & 4323 & 1081 & 5 & 2 \\
pol & 8065 & 2017 & 26 & 2 \\
predict\_students\_dropout\_and\_academic\_success & 3539 & 885 & 34 & 3 \\
qsar & 844 & 211 & 40 & 2 \\
rice\_cammeo\_and\_osmancik & 3048 & 762 & 7 & 2 \\
ringnorm & 5920 & 1480 & 20 & 2 \\
rl & 3976 & 994 & 12 & 2 \\
satimage & 5144 & 1286 & 36 & 6 \\
segment & 1848 & 462 & 17 & 7 \\
seismic+bumps & 2067 & 517 & 18 & 2 \\
shuttle & 46400 & 11600 & 9 & 7 \\
spambase & 3680 & 921 & 57 & 2 \\
splice & 2552 & 638 & 60 & 3 \\
sports\_articles\_for\_objectivity\_analysis & 800 & 200 & 59 & 2 \\
statlog & 800 & 200 & 20 & 2 \\
steel\_plates\_faults & 1552 & 389 & 27 & 7 \\
sylvine & 4099 & 1025 & 20 & 2 \\
taiwanese\_bankruptcy\_prediction & 5455 & 1364 & 95 & 2 \\
telco-customer-churn & 5634 & 1409 & 18 & 2 \\
thyroid & 5760 & 1440 & 21 & 3 \\
thyroid-ann & 3017 & 755 & 21 & 3 \\
thyroid-dis & 2240 & 560 & 26 & 5 \\
turiye\_student\_evaluation & 4656 & 1164 & 32 & 5 \\
twonorm & 5920 & 1480 & 20 & 2 \\
vehicle & 676 & 170 & 18 & 4 \\
wall-robot-navigation & 4364 & 1092 & 24 & 4 \\
water\_quality & 6396 & 1600 & 20 & 2 \\
waveform-5000 & 4000 & 1000 & 40 & 3 \\
waveform\_database\_generator & 3999 & 1000 & 21 & 3 \\
waveform\_database\_generator\_version\_1 & 4000 & 1000 & 21 & 3 \\
website\_phishing & 1082 & 271 & 9 & 3 \\
wine & 2043 & 511 & 4 & 2 \\
wine-quality-red & 1279 & 320 & 4 & 6 \\
wine-quality-white & 3918 & 980 & 11 & 7 \\
yeast & 1187 & 297 & 8 & 10 \\

\end{longtable}

\subsection{BCCO}
\label{app:data-bcco}

\begin{longtable}{p{0.52\textwidth}rrrr}
\caption{BCCO datasets used in the common 86-dataset evaluation. For each dataset, we report the effective context size, query size, number of input features, and number of classes.}
\label{tab:bcco_datasets} \\
\toprule
Dataset & Context Size & Query Size & \# Features & \# Classes \\
\midrule
\endfirsthead

\multicolumn{5}{c}%
{{\tablename\ \thetable{} -- continued from previous page}} \\
\toprule
Dataset & Context Size & Query Size & \# Features & \# Classes \\
\midrule
\endhead

\midrule
\multicolumn{5}{r}{{Continued on next page}} \\
\endfoot

\bottomrule
\endlastfoot

BNG(cmc) & 44236 & 11060 & 9 & 3 \\
CPMP-2015-runtime-classification & 368 & 159 & 22 & 4 \\
CostaMadre1 & 207 & 89 & 37 & 2 \\
Fitness\_Club\_c & 1200 & 300 & 6 & 2 \\
GAMETES\_Epistasis\_2-Way\_20atts\_0.1H\_EDM-1\_1 & 1120 & 480 & 20 & 2 \\
Gender\_Gap\_in\_Spanish\_WP & 3796 & 950 & 13 & 3 \\
GesturePhaseSegmentationProcessed & 6911 & 2962 & 32 & 5 \\
KDD & 4025 & 1007 & 45 & 2 \\
LED-display-domain-7digit & 350 & 150 & 7 & 10 \\
MIC & 1319 & 330 & 104 & 2 \\
Marketing\_Campaign & 1792 & 448 & 27 & 2 \\
National\_Health\_and\_Nutrition\_Health\_Survey & 1822 & 456 & 7 & 2 \\
PieChart1 & 493 & 212 & 37 & 2 \\
SPECTF & 244 & 105 & 44 & 2 \\
ad-click-data & 700 & 300 & 3 & 2 \\
ad-click-prediction-dataset & 7000 & 3000 & 7 & 2 \\
ada & 2902 & 1245 & 48 & 2 \\
airlines\_seed\_0\_nrows\_2000\_nclasses\_10\_ncols\_100\_stratify\_True & 1600 & 400 & 7 & 2 \\
amazon-survey-coupon-recommendation-dataset & 8878 & 3806 & 25 & 2 \\
analcatdata\_draft & 256 & 110 & 4 & 2 \\
arsenic-female-bladder & 391 & 168 & 4 & 2 \\
autoMpg & 278 & 120 & 7 & 2 \\
autoUniv-au4-2500 & 2000 & 500 & 100 & 3 \\
autoUniv-au7-1100 & 880 & 220 & 12 & 5 \\
balance-scale & 437 & 188 & 4 & 3 \\
bankadditionalfullcsv & 28831 & 12357 & 20 & 2 \\
baseball & 1072 & 268 & 16 & 3 \\
chemical-x & 1231 & 528 & 10 & 2 \\
chscase\_funds & 129 & 56 & 2 & 2 \\
chscase\_vine2 & 327 & 141 & 2 & 2 \\
classification-in-asteroseismology & 700 & 301 & 1 & 2 \\
colic & 257 & 111 & 26 & 2 \\
companion-plants & 696 & 299 & 3 & 4 \\
company\_bankruptcy\_prediction & 5455 & 1364 & 95 & 2 \\
compass & 13315 & 3329 & 17 & 2 \\
consultant-job-placement-data-2023 & 307 & 132 & 8 & 2 \\
coupons & 105 & 45 & 5 & 3 \\
customer-segmentation & 8068 & 2627 & 9 & 4 \\
dis & 3017 & 755 & 29 & 2 \\
disclosure\_x\_noise & 463 & 199 & 3 & 2 \\
employees-attrition-analysis & 3087 & 1323 & 27 & 2 \\
eye\_movements\_bin & 6086 & 1522 & 20 & 2 \\
fictional-character-battle-outcome-prediction & 1645 & 706 & 7 & 2 \\
first-order-theorem-proving & 4282 & 1836 & 51 & 6 \\
glioma\_grading\_clinical\_and\_mutation\_features\_dataset & 587 & 252 & 23 & 2 \\
golf\_play\_dataset\_extended & 876 & 219 & 9 & 2 \\
guitar-chord-finger-positioning & 205 & 89 & 9 & 7 \\
hcc-survival-data-set & 115 & 50 & 30 & 2 \\
heart-long-beach & 140 & 60 & 13 & 5 \\
hill-valley & 969 & 243 & 100 & 2 \\
income & 22384 & 9594 & 12 & 2 \\
jasmine & 2088 & 896 & 144 & 2 \\
jm1 & 8708 & 2177 & 21 & 2 \\
liver-disorders & 408 & 175 & 10 & 2 \\
loan-application-data & 350 & 150 & 11 & 2 \\
machine-learning & 14000 & 6000 & 12 & 2 \\
malware-analysis-datasets-pe-section-headers & 30305 & 12988 & 4 & 2 \\
maternal-health-risk & 709 & 305 & 6 & 3 \\
medium-app-reviews-from-google-play-store & 42886 & 18380 & 7 & 4 \\
mfeat-morphological & 1600 & 400 & 6 & 10 \\
mfeat-zernike & 1600 & 400 & 47 & 10 \\
microaggregation2 & 16000 & 4000 & 20 & 5 \\
nfl-combine-performance-data-2009-2019 & 2433 & 1044 & 15 & 2 \\
orange-quality & 168 & 73 & 10 & 8 \\
pbc & 292 & 126 & 18 & 2 \\
pc1 & 887 & 222 & 21 & 2 \\
phoneme & 3782 & 1622 & 5 & 2 \\
pm10 & 350 & 150 & 7 & 2 \\
predict-customer-purchase-behavior-dataset & 1050 & 450 & 6 & 2 \\
predict\_students\_dropout\_and\_academic\_success & 3539 & 885 & 34 & 3 \\
prnn\_synth & 175 & 75 & 2 & 2 \\
qsar-bioconcentration-classes-data-set & 545 & 234 & 11 & 3 \\
retail-chain-salespeople-engagement & 1466 & 629 & 64 & 2 \\
sentiment-analysis & 349 & 150 & 5 & 3 \\
simple-diabetes-dataset & 537 & 231 & 8 & 2 \\
sonar & 145 & 63 & 60 & 2 \\
spambase & 3680 & 921 & 57 & 2 \\
statlog & 800 & 200 & 20 & 2 \\
tabletpc-priceclassification & 1400 & 600 & 19 & 4 \\
tokyo1 & 671 & 288 & 44 & 2 \\
turiye\_student\_evaluation & 4656 & 1164 & 32 & 5 \\
user-knowledge & 282 & 121 & 5 & 5 \\
waveform-5000 & 4000 & 1000 & 40 & 3 \\
wholesale-customers & 308 & 132 & 8 & 2 \\
wine-quality-dataset-balanced-classification & 14700 & 6300 & 11 & 7 \\
wine-quality-red & 1279 & 320 & 4 & 6 \\

\end{longtable}

\end{document}